\documentclass[11pt]{article}

\usepackage[]{acl}

\usepackage{times}
\usepackage{caption}
\usepackage{latexsym}
\usepackage{diagbox} 
\usepackage{amsmath}   
\usepackage{amssymb}   
\usepackage{bm}        
\usepackage[T1]{fontenc}
\usepackage[utf8]{inputenc}
\usepackage{microtype}
\usepackage{inconsolata}
\usepackage{graphicx}
\usepackage[most]{tcolorbox}
\usepackage{fontawesome}
\usepackage[table]{xcolor}
\usepackage{times}
\usepackage{enumitem}
\usepackage{multirow}
\usepackage{placeins}
\usepackage{subcaption}
\usepackage{rotating} 
\usepackage{makecell}
\usepackage{algorithm}
\usepackage{algpseudocode}

\setlist{nosep}
\usepackage{soul}
\usepackage{colortbl}   
\usepackage{pgf}  
\usepackage{multicol}

\usepackage{dblfloatfix}
\usepackage{titlesec}
\usepackage{adjustbox}
\usepackage{booktabs}
\usepackage{dblfloatfix}    
\usepackage[most]{tcolorbox}
\usepackage{microtype}
\usepackage{array}
\usepackage{tabularx}
\usepackage{booktabs}
\usepackage{adjustbox}
\usepackage{ragged2e}
\usepackage{soul}
\usepackage{longtable}
\usepackage{titlesec}

\titlespacing*{\paragraph}
{0pt}    
{0.8ex}  
{0.8ex}  

\newcolumntype{L}[1]{>{\raggedright\arraybackslash}p{#1}}

\definecolor{lightgray}{gray}{0.5}
\definecolor{green}{rgb}{0.0, 1.0, 0.0}
\definecolor{myblue}{rgb}{0.0, 0.0, 1.0}   
\definecolor{lightblue}{rgb}{0.3, 0.5, 1.0}  

\definecolor{navyblue}{rgb}{0.0, 0.0, 0.5}
\definecolor{darkred}{rgb}{0.5, 0.0, 0.0} 
\newcommand{\HighSim}[1]{\colorbox{green!20}{\strut\text{#1}}}
\newcommand{\SomeSim}[1]{\colorbox{orange!25}{\strut\text{#1}}}
\newcommand{\MargSim}[1]{\colorbox{yellow!30}{\strut\text{#1}}}

\newcommand{\HighSimText}[1]{%
  {\sethlcolor{green!20}\hl{#1}}%
}

\newcommand{\SomeSimText}[1]{%
  {\sethlcolor{orange!25}\hl{#1}}%
}

\newcommand{\MargSimText}[1]{%
  {\sethlcolor{yellow!30}\hl{#1}}%
}

\sethlcolor{yellow!30}

\title{The Critical Role of Aspects in Measuring Document Similarity}
\author{
 \textbf{Eftekhar Hossain\textsuperscript{}},
 \textbf{Tarnika Hazra\textsuperscript{}},
 \textbf{Ahatesham Bhuiyan\textsuperscript{}},
 \textbf{Santu Karmaker\textsuperscript{}}
\\
 \textsuperscript{}Bridge-AI Lab@UCF, Department of Computer Science
\\University of Central Florida, USA
\\
 \small{
   \textbf{Correspondence:} \href{mailto:eftekhar@ucf.edu}{eftekhar@ucf.edu}, \href{santu@ucf.edu}{santu@ucf.edu}
 }
}

\begin{document}
\maketitle

\begin{abstract}



We introduce \textsc{AspectSim}, a \textit{simple} and  \textit{interpretable} framework that requires conditioning document similarity on an explicitly specified aspect, which is different from the traditional holistic approach in measuring document similarity. Experimenting with a newly constructed benchmark of {$\mathbf{26K}$} aspect–document pairs, we found that \textsc{AspectSim}, when implemented with direct GPT-4o prompting, achieves substantially higher \textit{human-machine agreement} ($\approx$\textbf{80\% higher}) than the same for holistic similarity without explicit aspects. These findings underscore the importance of explicitly accounting for aspects when measuring document similarity and highlight the need to revise standard practice. Next, we conducted a large-scale meta-evaluation using $16$ smaller open-source LLMs and $9$ embedding models with a focus on making \textsc{AspectSim} \textit{accessible} and \textit{reproducible}. While directly prompting LLMs to produce \textsc{AspectSim} scores turned out be ineffective (20-30\% \textit{human-machine agreement}), a simple two-stage refinement improved their agreement by $\approx$\textbf{140\%}. Nevertheless, agreement remains well below that of GPT-4o–based models, indicating that smaller open-source LLMs still lag behind large proprietary models in capturing aspect-conditioned similarity.

\end{abstract}



\section{Introduction}

Measuring document similarity is a core problem in NLP~\cite{morris2023text,chandrasekaran2021evolution}, underpinning applications such as document retrieval~\cite{guo2022semantic}, semantic matching~\cite{jiang2019semantic}, recommendation systems~\cite{contextualsimilarity}, etc. Most approaches in practice today estimate similarity using cosine distance in an embedding space, producing a single score that reflects overall semantic relatedness~\cite{cer2018universal,reimers2019sentence}, while others cast the problem as a binary classification task (similar vs.\ not similar).

\begin{figure}[!htb]
    \centering
    \includegraphics[width=0.93\linewidth]{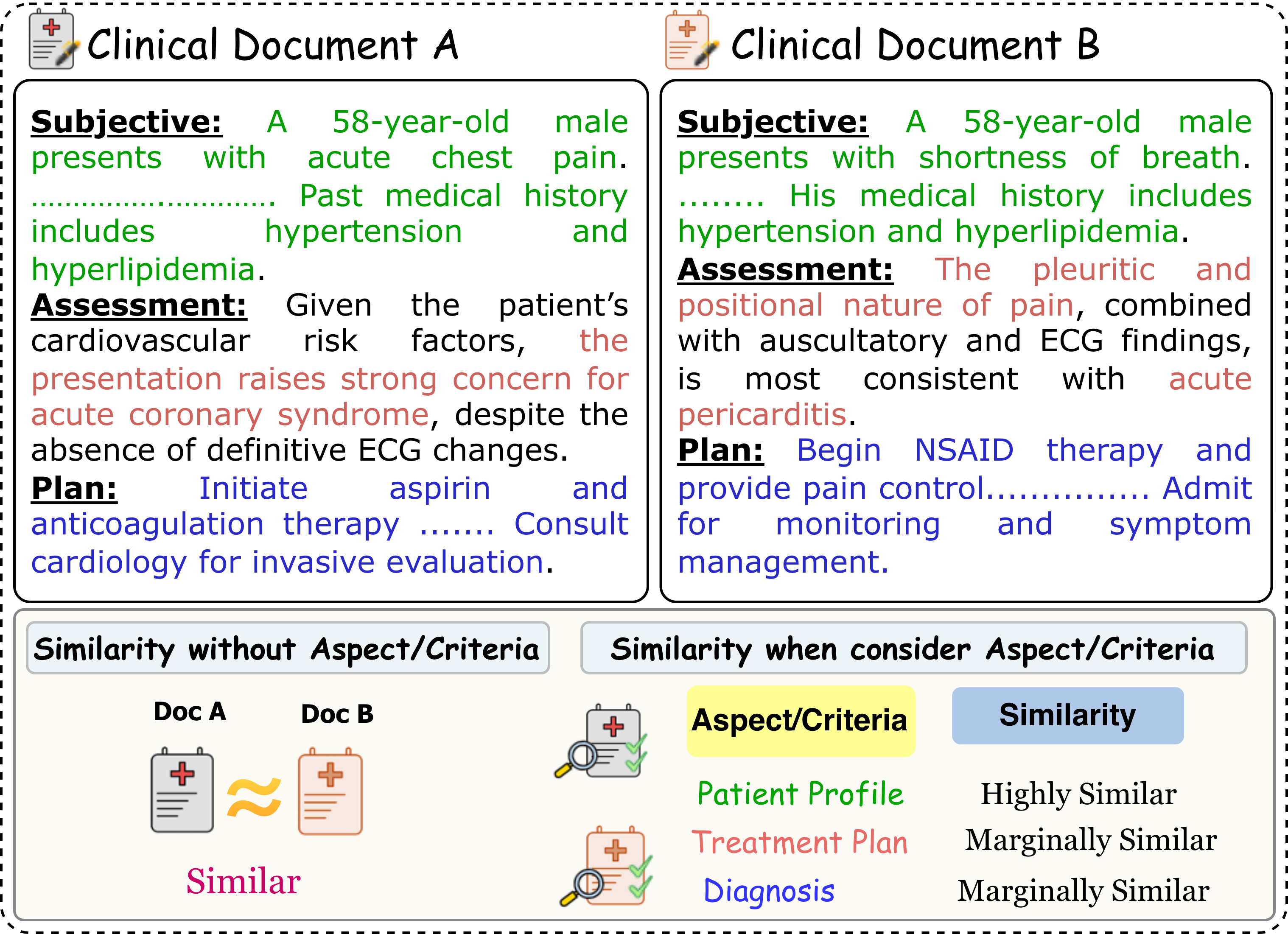}
    \vspace{-2mm}
    \caption{Illustration of how document similarity might differ when it is viewed through multiple aspects.}
    \vspace{-2mm}
\label{fig:paradiam}
\end{figure}

However, these practices oversimplify how humans assess document similarity, particularly for multi-sentence or long documents. Human judgments typically rely on multiple aspects or criteria implicitly defined by them, which can lead to substantial inter-annotator disagreement when these aspects differ. Similarly, cosine similarity over document embeddings suffers from analogous limitations as embedding dimensions are uninterpretable and sensitive to training variations, obscuring which aspects drive the resulting similarity scores. This lack of interpretability and control poses a practical bottleneck, as illustrated in Figure~\ref{fig:paradiam}, where two clinical notes for the same patient appear holistically similar yet differ markedly across specific aspects.

To address this gap, we introduce \textsc{AspectSim}, a framework for computing document similarity conditioned on an explicitly specified aspect. By making the comparison criterion explicit, \textsc{AspectSim} reduces ambiguity in similarity judgments, improves agreement among both human annotators and automated metrics, and enhances robustness in downstream NLP applications. Formally, \textsc{AspectSim} takes as input Document~A, Document~B, and a natural-language description of the aspect, and outputs an estimate of their aspect-based similarity.

Next, we examine the practical implementation of \textsc{AspectSim}, explore alternative design choices, and identify optimal configurations through a systematic study of key research questions. For experiments, we introduce a new benchmark of $\mathbf{26,000}$ instances spanning five domains (politics, factual events, scientific peer reviews, medical literature, and user-generated content), where each document pair is associated with an explicit rubric aspect and a graded similarity annotation. Using this benchmark, we begin with direct GPT-4o prompting as the simplest approach to computing \textsc{AspectSim} scores, and find it to be quite reliable, achieving approximately \textbf{90\% correlation} with human judgments (\textbf{+80\%} improvement over traditional holistic document similarity). However, reliance on proprietary models such as GPT-4o limits accessibility and reproducibility, both of which are essential for sustained scientific progress.

To make \textsc{AspectSim} accessible and reproducible, we investigate the use of smaller (1B-70B parameter range) open-source LLMs, such as Qwen-2.5-32B and LLaMA-3.3-70B, via direct prompting; however, this approach turned out to be ineffective, achieving only 20–30\% correlation with human judgments. We hypothesize that the resulting performance gap relative to GPT-4o arises from limitations in accurately extracting aspect-relevant evidence from document pairs. To mitigate this issue, we refine the \textsc{AspectSim} computation by decomposing it into two explicit stages: (i) extracting or retrieving content relevant to a given aspect, and (ii) computing embedding-based similarity over the extracted evidence. We then conduct a series of experiments to identify the optimal choices for 1) LLM used in aspect-based extraction and 2) embedding used in semantic matching.

A meta-evaluation across $16$ open-source LLMs and $9$ embedding models shows that the refined two-stage (\textit{extract--then--embed}) \textsc{AspectSim} substantially improves \textit{human–machine agreement} (\textbf{+140\%} improvement) compared to direct-LLM-prompting. We also found that relatively larger open-source models ($<70B$ parameters), e.g., Qwen-2.5-72B and LLaMA-3.3-70B, achieve the highest correlation with ground-truth labels ($\approx 58\%$) when used as extractors, while the choice of embedding model has only a marginal effect. Nevertheless, performance remains well below that of GPT-4o–based models, indicating that smaller open-source LLMs still lag behind large proprietary models in capturing aspect-conditioned semantics.

\section{Related Work}
Semantic similarity has been extensively studied in NLP. Before the deep learning era, document similarity was measured using symbolic and statistical methods such as TF–IDF–weighted bag-of-words representations, Jaccard similarity, and topic models (e.g., LSA, LDA). These approaches relied on lexical overlap or low-dimensional latent spaces. Current approaches use embedding-based similarity, such as the Universal Sentence Encoder~\cite{cer2018universal} and Sentence-BERT~\cite{reimers2019sentence}, which achieved strong correlation with human judgments on STS benchmarks~\cite{agirre2013sem} by modeling global semantic overlap. However, these methods ignore the conditional nature of similarity, where two documents may be considered ``similar'' under specific aspects, but judged as ``distinct'' otherwise.

\paragraph{Aspect-Based Document Similarity.}
Subsequent work explored aspect-based similarity, including aspect-conditioned comparisons of scientific papers~\cite{ostendorff2020aspect,ostendorff2022specialized}, multi-vector decompositions over fine-grained aspects~\cite{mysore2022multi}, contrastive learning of aspect-conditioned embeddings~\cite{schopf2023aspectcse}, and hierarchical aspect-based retrieval for scientific documents~\cite{wang2023scientific}. While these methods highlight the importance of aspect sensitivity, they typically rely on modifying the embedding space and operate over a small, predefined set of domain-specific aspects. In contrast, \textsc{AspectSim} enables \textbf{\textit{free-form, user-specified}} aspect conditioning without altering the embedding space, supporting arbitrary aspects across diverse domains.

\paragraph{Aspect-Guided Evidence Extraction.}

Aspect-conditioned similarity requires identifying document segments relevant to a target aspect. Prior work on explainable NLP has focused on extracting minimal segments as rationales for model predictions~\cite{lei2016rationalizing,deyoung2020eraser,thorne2018fever}. More recent studies leverage LLMs to localize semantically relevant evidence spans~\cite{zhao2024seer,peng2024metaie,barrow2025safepassage}, for example, in scientific information extraction~\cite{polak2024extracting} or clinical evidence identification~\cite{jiang2024evidence}. Unlike these approaches, which extract evidence to support predefined labels or entities, \textsc{AspectSim} supports the extraction of evidence based on any \textit{free-form user-defined aspect}.

\section{Method}\label{sec:method}

\subsection{Dataset Construction}

To our knowledge, no existing benchmark directly supports document similarity assessment conditioned on user-defined free-form aspects. Manually constructing such a large dataset is prohibitively costly and time-consuming. Recent work shows that LLMs can reliably act as data generators or judges for high-quality synthetic annotations~\cite{gptJudge1,gptJudge2,gptJudge3,gptJudge4,gptJudge5}. Motivated by this, we used GPT-4o to synthesize an aspect-conditioned similarity dataset as a scalable alternative to large-scale human annotation.

\subsubsection{Data Source Selection}
To curate the dataset, we sample document pairs with shared topical themes from five multi-document summarization corpora, which naturally contain semantically related texts centered on common events, arguments, or opinions, making them ideal sources for our task. Specifically, we drew samples from five widely used datasets: (i) Wikipedia Current Events Portal (WCEP/Wiki)~\cite{wcep}, (ii) Medical Studies Literature Review Summarization (MSLR)~\cite{ms2}, (iii) Argumentative Political Documents (AllSides/Side)~\cite{allside}, (iv) Scientific Peer Reviews (Peer)~\cite{peerreview}, and (v) User Opinion Summarization (OpinionSum/Hotel) \cite{opinionsum}. In total, we sampled approximately $2,100$ document pairs across these domains. The exact criteria for document sampling are provided in Appendix~\ref{sec:sampling}, and the final dataset statistics are provided in Table~\ref{tab:dataset_stats} of Appendix~\ref{appen:stat}.


\subsubsection{GPT-4o-Based Dataset Curation}\label{sec:curation}
For each document pair $(D_A, D_B)$, we instruct GPT-4o to perform a three-stage curation process designed to extract shared aspects and quantify similarity w.r.t. those aspects. The steps are as follows:


\paragraph{Task 1: Aspect Identification.}
Given two documents $(D_A, D_B)$, GPT-4o extracts a set of salient aspects $A = \{a_1, a_2, \dots, a_m\}$ that are discussed in both documents. Each aspect $a_i$ represents a salient topic or concept that occurs with sufficient contextual evidence in both documents.

\paragraph{Task 2: Aspect-Conditioned Extraction.}
For every identified aspect $a_i \in A$, the model extracts the corresponding segments/textual evidence $S_A^{(i)} \subseteq D_A, \quad S_B^{(i)} \subseteq D_B$ that are relevant to $a_i$. This evidence can be single or multiple sentences (span\footnote{A span refers to an extraction of more than one sentence.}) that contain the document’s viewpoint on the aspect.

\paragraph{Task 3: Silver Standard Ground Truth Generation.}
For each aspect $a_i$, GPT-4o compares $S_A^{(i)}$ against $S_B^{(i)}$ and assigns a similarity label $y^{(i)} \in \{\textit{Highly}, \textit{Somewhat}, \textit{Marginally}\}\text{ Similar}$. This is based on fine-grained semantic similarity defined in prior works on sentence and document-level semantics \cite{agirre2012semeval, cer2017semeval}.


\begin{itemize}[leftmargin=*]
    \item \textit{Highly Similar:} Evidence pair conveys nearly identical meaning, emphasis, and perspective.
    \item \textit{Somewhat Similar:} Partial semantic overlap on the target aspect, with differences in fine details.
    \item \textit{Marginally Similar:} Evidence pair diverges in meaning or stances despite a shared broad topic.
\end{itemize}


This structured process yields an aspect-aware document similarity dataset, where each sample consists of $(D_A, D_B, a_i, S_A^{(i)}, S_B^{(i)}, y^{(i)})$.

\subsubsection{Negative Instance Generation}
To ensure robust evaluation, we also include negative instances in the dataset in which certain aspects are absent from one of the two documents, allowing our metric to discern when a topic is truly unshared rather than merely weakly related. Specifically, for each document pair $(D_A, D_B)$, we prompt GPT-4o to identify aspects that appear exclusively in only one of the documents $A$ or $B$. If an aspect $a_j$ is found in $D_A$ but not in $D_B$ (or vice versa), we mark it as \textit{Not-Found}, denoted by $NF_B$ or $NF_A$. For detailed examples, see Appendix~\ref{appendix:prompts} and ~\ref{appendix:data-sample}.

\subsection{Data Quality Validation by Humans}
\label{sec:validation}

After dataset curation, we assessed the reliability of our GPT-4o–curated dataset by creating gold-standard ground-truth labels with two human annotators. To be more specific, we randomly sample $10\%$ of the instances from the GPT-4o-curated dataset ($2,536$ instances, comprising $446$ unique documents). Each instance is independently annotated by both annotators, who are also co-authors of this paper. Among the annotators, one has over 4 years of research experience in NLP and AI, while the other has a background in linguistics with extensive technical annotation expertise. Both annotators were instructed to assign the similarity labels based on the provided detailed annotation guidelines (Appendix~\ref{appendix:annot-aspect} and \ref{appendix:annot-aspect-free}).

For evaluation, we measured the agreement between each human annotator and GPT-4o labels using Cohen's $\kappa$ and found it to be quite high (overall $\kappa\approx 0.9$). Hence, we concluded that our \textit{GPT-4o-generated similarity labels were reliable}.


\subsection{Does Explicit Aspect Indeed Improve Similairy Assessment?}

To answer this question, we compared document similarity in both aspect-aware and aspect-agnostic settings to examine how the presence of an explicit aspect affects inter-annotator agreement in 2 cases: 1) agreement between two human annotators, and 2) agreement between a human and a GPT-4o-based validated annotator (see section~\ref{sec:validation} for validation details of GPT-4o).

\smallskip
\noindent\textbf{\textit{Human Vs Human:}} As summarized in Table~\ref{tab:human_agreement}, human annotators exhibit substantially higher agreement in the aspect-aware setting (overall $\kappa\approx 0.905$) than in the aspect-agnostic setting ($\kappa\approx 0.502$). This suggests that, in the absence of explicit aspect guidance, annotators likely rely on different implicit criteria when judging document similarity, resulting in lower agreement.

\smallskip
\noindent\textbf{\textit{Humans Vs GPT-4o:}} A similar pattern is observed between humans and GPT-4o annotator, whose agreement is consistently higher when aspects are provided ($\kappa\approx 0.9$) and degrades notably without aspect guidance ($\kappa = 0.46$).

\begin{table}[!htb]
\centering
\footnotesize
\begin{adjustbox}{width=0.95\linewidth}
\begin{tabular}{p{1.8cm} c c c c}
\toprule
\textbf{Dataset} & \textbf{\# Samples} & \textbf{IAA} & \textbf{A1 vs GPT} & \textbf{A2 vs GPT} \\
\midrule

\multicolumn{5}{l}{\textbf{Evaluation with Aspects}} \\
\midrule
AllSides & 836 & 0.927 & 0.912 & 0.925 \\
WCEP     & 739 & 0.920 & 0.901 & 0.916 \\
MS$^2$   & 608 & 0.881 & 0.937 & 0.925 \\
Peer     & 224 & 0.867 & 0.847 & 0.905 \\
Hotel    & 129 & 0.929 & 0.880 & 0.919 \\
\rowcolor{gray!15}
Overall  & 2,536 & \textbf{0.905} & \textbf{0.895} & \textbf{0.918} \\
\midrule

\multicolumn{5}{l}{\textbf{Evaluation without Aspects}} \\
\midrule
AllSides & 128 & 0.465 & 0.407 & 0.571 \\
WCEP     & 129 & 0.431 & 0.427 & 0.329 \\
MS$^2$   & 127 & 0.514 & 0.452 & 0.562 \\
Peer     & 38  & 0.539 & 0.445 & 0.516 \\
Hotel    & 24  & 0.565 & 0.600 & 0.400 \\
\rowcolor{gray!15}
Overall  & 446 & 0.502 & 0.466 & 0.475 \\
\bottomrule
\end{tabular}
\end{adjustbox}

\caption{Inter-annotator agreement (IAA) of human evaluation for document similarity under aspect-aware and aspect-agnostic settings, measured using Cohen’s $\kappa$. A1 and A2 denote two human annotators.}
\label{tab:human_agreement}
\end{table}

These results suggest that: 1) Spelling out the aspect during document similarity assessment indeed improves annotator agreement significantly, 2)  GPT-4o–generated annotations are reliable and can indeed be treated as a silver-standard ground truth for aspect-based similarity.

\subsection{Implementing \textsc{AspectSim} Metric}\label{sec:aspectsim}

Motivated by the improved agreement achieved through explicit aspect conditioning, we propose \textsc{AspectSim}, a metric that estimates the similarity between a document pair with respect to an explicitly defined aspect expressed in free-form natural language. Its implementation is straightforward: simply prompt GPT-4o to perform Tasks~2 and~3 in Section~\ref{sec:curation}, the resulting similarity scores directly yield reliable \textsc{AspectSim} estimates.

\section{Towards an Accessible and Reproducible \textsc{AspectSim} Metric}

Although a GPT-4o–prompting–based implementation of \textsc{AspectSim} is simple and effective, its proprietary nature and massive scale ($\sim$2T parameters) limit accessibility and reproducibility, both of which are essential for sustained scientific progress. This motivates a key question: \textit{Can aspect-aware similarity judgments be reliably assessed without proprietary LLMs?} To investigate this, we evaluate three open-source LLMs (Qwen-3-8B, Gemma-3-12B, and Phi-4-14B) as raters, using the same guidelines provided to human annotators (Section~\ref{sec:validation}). We assess reliability via inter-annotator agreement in two settings (Table~\ref{tab:llm_aggrement}): (i) agreement between pairs of LLM raters, and (ii) agreement between an LLM rater and our validated GPT-4o annotator.



\begin{table}[!htb]
\centering
\footnotesize
\begin{adjustbox}{max width=0.95\linewidth}
\begin{tabular}{p{1.6cm} c ccc}
\toprule

\textbf{Dataset} & \textbf{\# Samples}
& \textbf{IAA} & \textbf{L1 vs GPT} & \textbf{L2 vs GPT} \\
\midrule

\multicolumn{5}{l}{\textbf{Evaluation with Aspects}} \\
\midrule
AllSides & 836  & 0.195 & 0.104 & 0.065 \\
WCEP     & 739 & 0.172 & 0.406 & 0.296 \\
MS$^2$   & 608  & 0.207 & 0.397 & 0.236 \\
Peer     & 224  & 0.331 & 0.200 & 0.166 \\
Hotel    & 129 & 0.185 & 0.421 & 0.174 \\
\rowcolor{gray!15}
Overall & 2,536  & 0.218 & 0.306 & 0.187\\
\midrule
\multicolumn{5}{l}{\textbf{Evaluation without Aspects}} \\
\midrule
AllSides & 128 &  0.515 & 0.642 & 0.599 \\
WCEP     & 129 & 0.418 & 0.554 & 0.522 \\
MS$^2$   & 127  & 0.522 & 0.627 & 0.639 \\
Peer     & 38 & 0.493 & 0.610 & 0.597 \\
Hotel    & 24  & 0.501 & 0.516 & 0.432 \\
\rowcolor{gray!15}
Overall & 446 & 0.489 & 0.589 & 0.557 \\
\bottomrule
\end{tabular}
\end{adjustbox}

\vspace{-1mm}
\caption{Inter-annotator agreement (IAA) of LLM-based evaluation for document similarity under aspect-aware and aspect-agnostic settings, measured using Cohen’s $\kappa$. L1 and L2 denote the top two best LLM evaluators, Qwen-3-8B and Phi-4-14B, respectively.} 
\label{tab:llm_aggrement}\vspace{-2mm}
\end{table}


\subsection{Agreement Analysis}
\noindent\textbf{\textit{LLM Vs LLM:}} In contrast to human evaluation, agreement between LLMs is substantially lower in the aspect-aware setting ($\kappa\approx 0.218$) than in the aspect-agnostic setting ($\kappa\approx 0.489$), as shown in Table~\ref{tab:llm_aggrement}. This opposite trend indicates that while smaller LLMs can form relatively consistent judgments for holistic similarity, they struggle when similarity requires isolating aspect-specific content, leading to divergent implicit evidence selection and, consequently, inconsistent judgments.


\smallskip
\noindent\textbf{\textit{LLM Vs GPT-4o:}}  A similar degradation is observed in agreement between smaller LLMs and GPT-4o under aspect-aware evaluation ($\kappa\approx 0.25$), which highlights a key limitation of smaller LLMs.



\subsection{Improving \textsc{AspectSim} with Small LLMs}
Although these results are disappointing, they raise an important research question: How can the performance of smaller open-source LLMs be improved for aspect-aware similarity assessment? We hypothesize that the performance gap relative to GPT-4o stems from limitations in accurately extracting aspect-relevant evidence from document pairs. To address this, we refine the \textsc{AspectSim} computation by decomposing it into two explicit steps (Algorithm~\ref{alg:aspectsim}): (i) extracting or retrieving content relevant to a given aspect, and (ii) computing embedding-based similarity over the extracted evidence. We argue that this separation will enable smaller LLMs to focus on each subtask more effectively, resulting in improved overall performance. More details can be found in Appendix~\ref{sec:aspetsim}.  



\begin{algorithm}[!htb]
\footnotesize

\begin{algorithmic}[1]
\Require Document pair $(D_A, D_B)$, aspect $a$
\Ensure Aspect-aware similarity score $s$

\State $S_A \gets \textsc{Extract}(D_A, a)$ \Comment{extract relevant content}
\State $S_B \gets \textsc{Extract}(D_B, a)$
\State $v_A \gets \textsc{Embed}(S_A)$ \Comment{compute embeddings}
\State $v_B \gets \textsc{Embed}(S_B)$
\State $s \gets \textsc{CosineSimilarity}(v_A, v_B)$
\State \Return $s$
\end{algorithmic}

\caption{Pseudocode for \textsc{AspectSim}}
\label{alg:aspectsim}
\end{algorithm}
A robust implementation of Algorithm~\ref{alg:aspectsim} critically depends on the effective realization of its \textsc{Extract} and \textsc{Embed} methods. Accordingly, we next evaluate which smaller open source language models are most suitable for implementing these functions, with the goal of developing an accessible and reproducible \textsc{AspectSim} framework.

\section{Experiments with Smaller LLMs}\label{sec:experiments}

\vspace{-2mm}
\subsection{Small Open Source Models}
For \textsc{AspectSim}, we tested a diverse set of 16 (relatively smaller) open-source LLMs spanning six prominent model families: Mistral~\cite{jiang2023mistral7b}, LLaMA~\cite{dubey2024llama}, Phi~\cite{abdin2024phi}, Gemma~\cite{team2024gemma}, Qwen~\cite{yang2025qwen3}, and DeepSeek-R1~\cite{guo2025deepseek} covering a broad range of parameter scales. Specifically, the models include: Mistral-7B; LLaMA-3.2 (1B, 3B) and LLaMA-3.3-70B; Gemma-2-27B and Gemma-3 (1B, 12B); Phi-4 (4B, 14B); Qwen-2.5 (14B, 32B, 72B); Qwen-3 (4B, 8B); and distilled Qwen variants of DeepSeek-R1 (14B, 32B). For embeddings, we tested nine models selected based on their top performance on the MTEB\footnote{https://huggingface.co/spaces/mteb/leaderboard} Semantic Textual Similarity (STS) leaderboard. The models include MPNet~\cite{song2020mpnet}, Jina-V3~\cite{sturua2024jina}, Bilingual~\cite{bilingual}, Mxbai~\cite{emb2024mxbai}, SFR (S)-Mistral~\cite{SFRAIResearch2024}, Ling (L)-Mistral~\cite{LinqAIResearch2024}, E5~\cite{e5}, Embedding from Qwen-3~\cite{qwen3embedding}, and EmbeddingGemma~\cite{vera2025embeddinggemma}.

\subsection{Baseline Methods}
We compare \textsc{AspectSim} metric against three aspect-aware baseline approaches: 

\smallskip\noindent \textbf{LLM-Based Similarity (LBS). }
In this setting, we directly prompt LLMs to estimate the similarity between two documents with respect to a given aspect, producing a score in the range $[0,1]$.

\smallskip\noindent
\textbf{Whole Document Similarity (WDS). }
This baseline represents the simplest embedding-based comparison, computing cosine similarity between document embeddings without aspect conditioning. The resulting similarity scores are then correlated with the ground-truth aspect similarity labels.

\smallskip\noindent
\textbf{Projection Similarity Difference (PSD). }
The third baseline extends standard embedding similarity by incorporating aspect-conditioned projections. Each document embedding is projected along the direction of the aspect embedding, and the absolute difference between the projections is used as the similarity score, which is then correlated with the ground-truth labels. Exact formulations are provided in Appendix~\ref{appendix:psd}.

\begin{table*}[!htb]

\tiny
  \centering
  
\resizebox{0.9\textwidth}{!}{ 

      \begin{tabular}{p{1.5cm}||ccccccccc||cc} 
      
        \toprule
         \multicolumn{12}{c}{\textbf{Retrieve-Then-Embed (Sentence Level)}} \\ 
        \cmidrule(r){1-11} \cmidrule(l){12-12}
         \diagbox[width=1.5cm, height=0.6cm]{LLMs}{Emb.} &MPNet &Jinja &Bilingual & Mxbai & S-Mistral & L-Mistral & E5 & Qwen3 & Gemma & Avg. \\
        \midrule
        Mistral-7B &  \cellcolor{myblue!7}0.501	&  \cellcolor{myblue!7}0.502	&  \cellcolor{myblue!7}0.519	&  \cellcolor{myblue!7}0.512	&  \cellcolor{myblue!8}0.523	&  \cellcolor{myblue!8}0.527 &  \cellcolor{myblue!8}0.522	&  \cellcolor{myblue!8}0.523 &  \cellcolor{myblue!8}0.516	& 0.516 \\
        LLaMA-3.2-1B & \cellcolor{darkred!15}0.097 & \cellcolor{darkred!15}0.109 & \cellcolor{darkred!15}0.107  & \cellcolor{darkred!15}0.105  &  \cellcolor{darkred!15}0.111  & 	\cellcolor{darkred!15}0.109  & \cellcolor{darkred!15}0.104  & \cellcolor{darkred!15}0.118 & \cellcolor{darkred!15}0.111 & 0.107 \\
        LLaMA-3.2-3B &  \cellcolor{myblue!2}0.444 &  \cellcolor{myblue!3}0.455 &  \cellcolor{myblue!3}0.466 &  \cellcolor{myblue!3}0.462 &  \cellcolor{myblue!3}0.469 &  \cellcolor{myblue!3}0.475 &  \cellcolor{myblue!3}0.466 & \cellcolor{myblue!3} 0.471 &  \cellcolor{myblue!3}0.466 & 0.464  \\
        LLaMA-3.3-70B &  \cellcolor{myblue!11}0.575 &  \cellcolor{myblue!11}0.576 &  \cellcolor{myblue!25}0.591 &  \cellcolor{myblue!18}0.585 &  \cellcolor{myblue!25}0.591 &  \cellcolor{myblue!25}\textbf{0.596} &  \cellcolor{myblue!25}0.592 &  \cellcolor{myblue!25}{0.591} & \cellcolor{myblue!18}0.588 & \textbf{0.587}  \\
        Gemma-3-1B & \cellcolor{darkred!15}0.092 & \cellcolor{darkred!15}0.095 & \cellcolor{darkred!15}0.113 & \cellcolor{darkred!15}0.107 & \cellcolor{darkred!15}0.109	 & \cellcolor{darkred!15}0.107 & \cellcolor{darkred!15}0.105 & \cellcolor{darkred!15}0.108 & \cellcolor{darkred!15}0.102 & 0.104 \\
        Gemma-3-12B &  \cellcolor{myblue!8}0.531	&  \cellcolor{myblue!8}0.534 &  \cellcolor{myblue!9}0.555&  \cellcolor{myblue!8}0.547 &  \cellcolor{myblue!9}0.555 &  \cellcolor{myblue!10}0.561 &  \cellcolor{myblue!9}0.556 &  \cellcolor{myblue!9}0.554 &  \cellcolor{myblue!9}0.552 & 0.549  \\
        Gemma-2-27B &  \cellcolor{myblue!8}0.549	 &  \cellcolor{myblue!8}0.549 &  \cellcolor{myblue!11}0.574 &  \cellcolor{myblue!10}0.566 & 	 \cellcolor{myblue!10}0.568 &  \cellcolor{myblue!11}0.573 &  \cellcolor{myblue!11}0.572 &  \cellcolor{myblue!11}0.571 &  \cellcolor{myblue!10}0.567 & 0.566  \\
        Phi-4-4B &  \cellcolor{myblue!4}0.475 & \cellcolor{myblue!4}0.476 &  \cellcolor{myblue!5}0.487 &	 \cellcolor{myblue!5}0.482 &  \cellcolor{myblue!5}0.483 &  \cellcolor{myblue!5}0.484 &  \cellcolor{myblue!5}0.485 &  \cellcolor{myblue!5}0.484 &  \cellcolor{myblue!5}0.483 & 0.482  \\
        Phi-4-14B &  \cellcolor{myblue!10}0.563	&  \cellcolor{myblue!10}0.563	 &  \cellcolor{myblue!18}0.580 &  \cellcolor{myblue!11}0.577 &  \cellcolor{myblue!18}0.580 &  \cellcolor{myblue!18}0.584 &  \cellcolor{myblue!18}0.580 &  \cellcolor{myblue!11}0.577 &  \cellcolor{myblue!11}0.575 & 0.576 \\
        Qwen-3-4B &  \cellcolor{myblue!9}0.557 &  \cellcolor{myblue!9}0.557 &  \cellcolor{myblue!11}0.570 &  \cellcolor{myblue!10}0.566 &  \cellcolor{myblue!10}0.566 &  \cellcolor{myblue!10}0.569 &   \cellcolor{myblue!11}0.570 &  \cellcolor{myblue!11}0.570 &  \cellcolor{myblue!10}0.566  & 0.566  \\
        Qwen-3-8B &  \cellcolor{myblue!11}0.571 &  \cellcolor{myblue!10}0.567 &  \cellcolor{myblue!18}0.583 &  \cellcolor{myblue!11}0.577 &  \cellcolor{myblue!18}0.581 &  \cellcolor{myblue!18}0.584 &  \cellcolor{myblue!18}0.582 &  \cellcolor{myblue!18}0.580 &  \cellcolor{myblue!11}0.577 & 0.578  \\
        Qwen-2.5-14B &  \cellcolor{myblue!10}0.566 & 	 \cellcolor{myblue!10}0.566 &  \cellcolor{myblue!11}0.579 &  \cellcolor{myblue!11}0.578 &  \cellcolor{myblue!11}0.577 &  \cellcolor{myblue!18}0.582 &  \cellcolor{myblue!11}0.579 &  \cellcolor{myblue!11}0.577 &  \cellcolor{myblue!11}0.576 & 0.576 \\
        Qwen-2.5-32B &  \cellcolor{myblue!11}0.576 & \cellcolor{myblue!11} 	0.576 &  \cellcolor{myblue!18}0.588 &  \cellcolor{myblue!18}0.586 &  \cellcolor{myblue!18}0.589 &  \cellcolor{myblue!25}0.594 &  \cellcolor{myblue!25}0.590 &  \cellcolor{myblue!18}0.586 &  \cellcolor{myblue!18}0.586 & 0.586 \\
        Qwen-2.5-72B &  \cellcolor{myblue!11}0.573 & 	 \cellcolor{myblue!11}0.572 &  \cellcolor{myblue!25}0.591 &   \cellcolor{myblue!18}0.586 & \cellcolor{myblue!18} 0.587 &  \cellcolor{myblue!25}0.591 &  \cellcolor{myblue!25}0.590 &  \cellcolor{myblue!18}0.588 &  \cellcolor{myblue!18}0.588 & 0.585  \\
        DeepSeek-14B &  \cellcolor{myblue!9}0.556 & 	 \cellcolor{myblue!9}0.552 &  \cellcolor{myblue!10}0.569 &  \cellcolor{myblue!10}0.568 &  \cellcolor{myblue!11}0.571 &  \cellcolor{myblue!11}0.574 &  \cellcolor{myblue!10}0.569 &  \cellcolor{myblue!10}0.567 &  \cellcolor{myblue!10}0.567 & 0.566  \\
        DeepSeek-32B & \cellcolor{myblue!9}0.553 & 	\cellcolor{myblue!9}0.554 & \cellcolor{myblue!11}0.571 &  \cellcolor{myblue!10}0.566 & \cellcolor{myblue!10}0.569 & \cellcolor{myblue!11}0.572 & \cellcolor{myblue!11}0.573 & \cellcolor{myblue!10}0.566 & \cellcolor{myblue!10}0.567 & 0.566  \\
        \midrule
        
       \multicolumn{12}{c}{\textbf{Retrieve-Then-Embed (Span Level)}}  \\ 
      \midrule
      Mistral-7B &  \cellcolor{myblue!7}0.517	&  \cellcolor{myblue!7}0.515	&  \cellcolor{myblue!8}0.535	&  \cellcolor{myblue!7}0.526	&  \cellcolor{myblue!7}0.528	&  \cellcolor{myblue!7}0.528 &  \cellcolor{myblue!7}0.529	&  \cellcolor{myblue!8}0.530 &  \cellcolor{myblue!8}0.533	& 0.527 \\

        LLaMA-3.2-1B & \cellcolor{darkred!15}0.087 & \cellcolor{darkred!15}0.096 & \cellcolor{darkred!15}0.102  & \cellcolor{darkred!15}0.103  &  \cellcolor{darkred!15}0.102  & 	\cellcolor{darkred!15}0.104  & \cellcolor{darkred!15}0.095  & \cellcolor{darkred!15}0.107 & \cellcolor{darkred!15}0.102 & 0.10  \\
        LLaMA-3.2-3B &  \cellcolor{myblue!2}0.444 &  \cellcolor{myblue!3}0.447 &  \cellcolor{myblue!3}0.460 &  \cellcolor{myblue!3}0.478 &  \cellcolor{myblue!3}0.465 &  \cellcolor{myblue!3}0.466 &  \cellcolor{myblue!3}0.466 & 
        \cellcolor{myblue!3} 0.471 &  \cellcolor{myblue!3}0.472 & 0.466  \\
        
        LLaMA-3.3-70B &   \cellcolor{myblue!10}0.569 &  \cellcolor{myblue!11}0.573 &  \cellcolor{myblue!18}0.586 &  \cellcolor{myblue!11}0.577 &  \cellcolor{myblue!11}0.578 &  \cellcolor{myblue!11}0.579 &  \cellcolor{myblue!18}0.581 & \cellcolor{myblue!18}0.586 & \cellcolor{myblue!18}0.584 & 0.579 \\
        
        Gemma-3-1B & \cellcolor{darkred!15}0.014 & \cellcolor{darkred!15}0.145 & \cellcolor{darkred!15}0.153 & \cellcolor{darkred!15}0.142 & \cellcolor{darkred!15}0.154	 & \cellcolor{darkred!15}0.153 & \cellcolor{darkred!15}0.152 & \cellcolor{darkred!15}0.137 & \cellcolor{darkred!15}0.146 & 0.146 \\
        
        Gemma-3-12B &  \cellcolor{myblue!8}0.542	&  \cellcolor{myblue!8}0.544 &  \cellcolor{myblue!9}0.561&  \cellcolor{myblue!8}0.548 &  \cellcolor{myblue!8}0.550 &  \cellcolor{myblue!8}0.550 &  \cellcolor{myblue!8}0.557 &  \cellcolor{myblue!9}0.560 &  \cellcolor{myblue!8}0.554 & 0.552 \\
        
        Gemma-2-27B &  \cellcolor{myblue!9}0.564	 &  \cellcolor{myblue!9}0.566 &  \cellcolor{myblue!18}0.581 &  \cellcolor{myblue!11}0.572 & 	 \cellcolor{myblue!11}0.571 &  \cellcolor{myblue!11}0.573 &  \cellcolor{myblue!11}0.573 &  \cellcolor{myblue!18}0.580 &  \cellcolor{myblue!18}0.579 & 0.573  \\
        
        Phi-4-4B &  \cellcolor{myblue!5}0.496 & \cellcolor{myblue!5}0.495 &  \cellcolor{myblue!5}0.504 &	 \cellcolor{myblue!5}0.504 &  \cellcolor{myblue!5}0.497 &  \cellcolor{myblue!5}0.498 &  \cellcolor{myblue!5}0.503 &  \cellcolor{myblue!5}0.497 &  \cellcolor{myblue!5}0.501 & 0.50 \\
        
        Phi-4-14B &  \cellcolor{myblue!10}0.568	&  \cellcolor{myblue!11}0.571	 &  \cellcolor{myblue!18}0.581 &  \cellcolor{myblue!11}0.573 &  \cellcolor{myblue!11}0.574 &  \cellcolor{myblue!11}0.576 &  \cellcolor{myblue!11}0.576 &  \cellcolor{myblue!18}0.581 &  \cellcolor{myblue!11}0.578 & 0.575 \\
        
        Qwen-3-4B &  \cellcolor{myblue!10}0.560 &  \cellcolor{myblue!9}0.561 &  \cellcolor{myblue!11}0.570 &  \cellcolor{myblue!10}0.564 &  \cellcolor{myblue!10}0.563 &  \cellcolor{myblue!10}0.564 &   \cellcolor{myblue!10}0.564 &  \cellcolor{myblue!11}0.571 &  \cellcolor{myblue!10}0.565  & 0.566  \\
        
        Qwen-3-8B &  \cellcolor{myblue!10}0.564 &  \cellcolor{myblue!10}0.562 &  \cellcolor{myblue!11}0.575 &  \cellcolor{myblue!11}0.569 &  \cellcolor{myblue!10}0.564 &  \cellcolor{myblue!10}0.566 &  \cellcolor{myblue!10}0.567 &  \cellcolor{myblue!11}0.572 &  \cellcolor{myblue!10}0.569 & 0.578 \\
        
        Qwen-2.5-14B &  \cellcolor{myblue!10}0.567 & 	 \cellcolor{myblue!11}0.570 &  \cellcolor{myblue!18}0.580 &  \cellcolor{myblue!11}0.575 &  \cellcolor{myblue!11}0.575 &  \cellcolor{myblue!11}0.577 &  \cellcolor{myblue!11}0.578 &  \cellcolor{myblue!18}0.581 &  \cellcolor{myblue!11}0.578 & 0.576 \\
        
        Qwen-2.5-32B &  \cellcolor{myblue!11}0.565 & \cellcolor{myblue!11} 	0.570 &  \cellcolor{myblue!18}0.582 &  \cellcolor{myblue!11}0.572 &  \cellcolor{myblue!11}0.573 &  \cellcolor{myblue!11}0.574 &  \cellcolor{myblue!11}0.574 &  \cellcolor{myblue!18}0.579 &  \cellcolor{myblue!11}0.576 & 0.574 \\
        
        Qwen-2.5-72B &  \cellcolor{myblue!11}0.574 & 	 \cellcolor{myblue!11}0.578 &  \cellcolor{myblue!18}0.588 &   \cellcolor{myblue!18}0.579 & \cellcolor{myblue!18} 0.583 &  \cellcolor{myblue!18}0.584 &  \cellcolor{myblue!18}0.584 &  \cellcolor{myblue!25}\textbf{0.590} &  \cellcolor{myblue!18}0.585 & \textbf{0.583}  \\
        
        DeepSeek-14B &  \cellcolor{myblue!10}0.561 & 	 \cellcolor{myblue!9}0.560 &  \cellcolor{myblue!11}0.575 &  \cellcolor{myblue!10}0.564 &  \cellcolor{myblue!10}0.565 &  \cellcolor{myblue!11}0.566 &  \cellcolor{myblue!10}0.568 &  \cellcolor{myblue!10}0.569 &  \cellcolor{myblue!10}0.568 & 0.565 \\
        
        DeepSeek-32B & \cellcolor{myblue!9}0.551 & 	\cellcolor{myblue!9}0.552 & \cellcolor{myblue!10}0.568 &  \cellcolor{myblue!10}0.560 & \cellcolor{myblue!10}0.562 & \cellcolor{myblue!10}0.564 & \cellcolor{myblue!10}0.566 & \cellcolor{myblue!10}0.566 & \cellcolor{myblue!10}0.563 & 0.566 \\
\midrule
     \multicolumn{12}{c}{\textbf{Summarize-Then-Embed}}  \\ 
      \midrule
      Mistral-7B &   \cellcolor{myblue!2}0.455 & \cellcolor{myblue!2}0.454 & \cellcolor{myblue!3}0.471 & \cellcolor{myblue!2}0.463 &
      \cellcolor{myblue!3}0.488 & \cellcolor{myblue!3}0.483 &
      \cellcolor{myblue!2}0.467 & \cellcolor{myblue!3}0.476 &
      \cellcolor{myblue!2}0.466 & 0.469\\

        LLaMA-3.2-1B & \cellcolor{darkred!15}0.039 & \cellcolor{darkred!15}0.058 &
        \cellcolor{darkred!15}0.058 & \cellcolor{darkred!15}0.043 &
        \cellcolor{darkred!15}0.059 & \cellcolor{darkred!15}0.054 &
        \cellcolor{darkred!15}0.048 & \cellcolor{darkred!15}0.069 &
        \cellcolor{darkred!15}0.061 & 0.054 \\
        
        LLaMA-3.2-3B &  \cellcolor{myblue!2}0.388 & \cellcolor{myblue!2}0.390 &
        \cellcolor{myblue!2}0.402 & \cellcolor{myblue!2}0.396 &
        \cellcolor{myblue!3}0.406 & \cellcolor{myblue!3}0.404 &
        \cellcolor{myblue!2}0.397 & \cellcolor{myblue!2}0.395 &
        \cellcolor{myblue!2}0.395 & 0.397  \\
        
        LLaMA-3.3-70B &  \cellcolor{myblue!7}0.530 & \cellcolor{myblue!7}0.528 &
        \cellcolor{myblue!9}0.544 & \cellcolor{myblue!8}0.537 &
        \cellcolor{myblue!10}0.551 & \cellcolor{myblue!9}0.549 &
        \cellcolor{myblue!8}0.540 & \cellcolor{myblue!9}0.544 &
        \cellcolor{myblue!8}0.542 & 0.541 \\
        
        Gemma-3-1B & \cellcolor{darkred!15}0.068 & \cellcolor{darkred!15}0.055 &
        \cellcolor{darkred!15}0.061 & \cellcolor{darkred!15}0.056 &
        \cellcolor{darkred!15}0.071 & \cellcolor{darkred!15}0.069 &
        \cellcolor{darkred!15}0.045 & \cellcolor{darkred!15}0.051 &
        \cellcolor{darkred!15}0.045 & 0.058\\
        
        Gemma-3-12B &
        \cellcolor{myblue!6}0.511 & \cellcolor{myblue!6}0.511 &
        \cellcolor{myblue!8}0.528 & \cellcolor{myblue!7}0.520 &
        \cellcolor{myblue!9}0.543 & \cellcolor{myblue!9}0.542 &
        \cellcolor{myblue!7}0.524 & \cellcolor{myblue!8}0.535 &
        \cellcolor{myblue!8}0.528 & 0.527 \\
        
        Gemma-2-27B &
        \cellcolor{myblue!3}0.499 & \cellcolor{myblue!5}0.504 &
        \cellcolor{myblue!8}0.523 & \cellcolor{myblue!7}0.512 &
        \cellcolor{myblue!9}0.531 & \cellcolor{myblue!9}0.531 &
        \cellcolor{myblue!7}0.513 & \cellcolor{myblue!9}0.532 &
        \cellcolor{myblue!8}0.521 & 0.518 \\
        
        Phi-4-4B &
        \cellcolor{myblue!2}0.349 & \cellcolor{myblue!2}0.351 &
        \cellcolor{myblue!2}0.355 & \cellcolor{myblue!2}0.354 &
        \cellcolor{myblue!2}0.360 & \cellcolor{myblue!2}0.360 &
        \cellcolor{myblue!2}0.356 & \cellcolor{myblue!2}0.352 &
        \cellcolor{myblue!2}0.355 & 0.355 \\
        
        Phi-4-14B &
        \cellcolor{myblue!10}0.557 & \cellcolor{myblue!10}0.556 &
        \cellcolor{myblue!11}0.569 & \cellcolor{myblue!10}0.565 &
        \cellcolor{myblue!18}0.577 & \cellcolor{myblue!18}0.576 &
        \cellcolor{myblue!11}0.567 & \cellcolor{myblue!18}0.572 &
        \cellcolor{myblue!10}0.565 & 0.567 \\
        
        Qwen-3-4B &
        \cellcolor{myblue!6}0.514 & \cellcolor{myblue!6}0.512 &
        \cellcolor{myblue!7}0.523 & \cellcolor{myblue!7}0.520 &
        \cellcolor{myblue!8}0.535 & \cellcolor{myblue!8}0.531 &
        \cellcolor{myblue!7}0.525 & \cellcolor{myblue!7}0.527 &
        \cellcolor{myblue!6}0.519 & 0.523 \\
        
        Qwen-3-8B &
        \cellcolor{myblue!9}0.548 & \cellcolor{myblue!9}0.546 &
        \cellcolor{myblue!10}0.557 & \cellcolor{myblue!10}0.552 &
        \cellcolor{myblue!11}0.561 & \cellcolor{myblue!11}0.560 &
        \cellcolor{myblue!10}0.557 & \cellcolor{myblue!10}0.558 &
        \cellcolor{myblue!10}0.555 & 0.555 \\
        
        Qwen-2.5-14B &
        \cellcolor{myblue!10}0.551 & \cellcolor{myblue!9}0.548 &
        \cellcolor{myblue!11}0.562 & \cellcolor{myblue!10}0.557 &
        \cellcolor{myblue!18}0.571 & \cellcolor{myblue!11}0.568 &
        \cellcolor{myblue!11}0.564 & \cellcolor{myblue!11}0.562 &
        \cellcolor{myblue!11}0.560 & 0.560 \\

        Qwen-2.5-32B &
        \cellcolor{myblue!9}0.547 & \cellcolor{myblue!9}0.546 &
        \cellcolor{myblue!11}0.562 & \cellcolor{myblue!10}0.553 &
        \cellcolor{myblue!18}0.571 & \cellcolor{myblue!11}0.567 &
        \cellcolor{myblue!10}0.558 & \cellcolor{myblue!11}0.565 &
        \cellcolor{myblue!10}0.559 & 0.559  \\
        
        Qwen-2.5-72B &
        \cellcolor{myblue!11}0.561 & \cellcolor{myblue!11}0.563 &
        \cellcolor{myblue!18}0.574 & \cellcolor{myblue!11}0.568 &
        \cellcolor{myblue!18}\textbf{0.578} & \cellcolor{myblue!18}0.576 &
        \cellcolor{myblue!18}0.571 & \cellcolor{myblue!18}0.573 &
        \cellcolor{myblue!18}0.572 & \textbf{0.571} \\

        DeepSeek-14B &
        \cellcolor{myblue!2}0.467 & \cellcolor{myblue!2}0.467 &
        \cellcolor{myblue!3}0.480 & \cellcolor{myblue!2}0.475 &
        \cellcolor{myblue!3}0.487 & \cellcolor{myblue!3}0.485 &
        \cellcolor{myblue!3}0.480 & \cellcolor{myblue!3}0.482 &
        \cellcolor{myblue!2}0.479 & 0.478  \\

        DeepSeek-32B &
        \cellcolor{myblue!9}0.546 & \cellcolor{myblue!9}0.543 &
        \cellcolor{myblue!10}0.556 & \cellcolor{myblue!10}0.551 &
        \cellcolor{myblue!11}0.562 & \cellcolor{myblue!10}0.559 &
        \cellcolor{myblue!10}0.557 & \cellcolor{myblue!10}0.555 &
        \cellcolor{myblue!10}0.554 & 0.554 \\

        \bottomrule
      \end{tabular}
}

    \caption{Performance of \textsc{AspectSim} across three design variations, measured by Spearman correlation over nine embedding models. Darker blue indicates higher correlations, while orange denotes lower values. Sentence-level extracts the most relevant sentence, span-level aggregates multiple relevant sentences, and summarize-then-embed applies aspect-conditioned summarization before computing similarity.} 
\vspace{-4mm}
  \label{tab:main-results}
\end{table*}

\subsection{Results}\label{sec:results}

\subsubsection{Compaing \textsc{AspectSim} Variants}
Table~\ref{tab:main-results} reports the performance of three variations of \textsc{AspectSim}: \textit{\textbf{Retrieve-Then-Embed (sentence-level)}}, which retrieves the single most representative sentence before embedding; \textit{R\textbf{etrieve-Then-Embed (span-level)}}, which retrieves multiple relevant sentences; and \textit{\textbf{Summarize-Then-Embed}}, which summarizes the aspect-relevant content before embedding for similarity estimation. The following trends are observed:

\noindent
\textbf{Scaling and Design effects.} 
Overall, \textsc{AspectSim} performance improves with model scale, though not monotonically. Very small models (1--3B) consistently underperform, rarely exceeding correlations of 0.45 against GPT-4o labels, reflecting limited capacity for aspect-relevant semantics. Mid-sized models (4--8B) show substantial gains ($\approx$0.50--0.57), with Qwen-3-8B performing particularly well. Models in the 12--14B range achieve strong performance ($\approx$0.56--0.58), capturing most scaling benefits, while larger models (27--72B) offer only marginal improvements, indicating diminishing returns. Across all scales, \textit{retrieve-then-embed} variants generally outperform \textit{summarize-then-embed} designs.



\smallskip\noindent
\textbf{Effect of embeddings.} Among the embedding variants, \textsc{E5}, \textsc{Qwen3}, and \textsc{Gemma} achieve the highest and most consistent correlations. \textsc{MPNet} and \textsc{Mxbai} perform competitively but remain mid-tier, whereas \textsc{Jina} and \textsc{Mistral} lag behind, suggesting limited aspect alignment capacity.

\subsubsection{Comparison with Baselines.}


\noindent
\textbf{Does removing aspect-based retrieval affect similarity estimation performance?}
To isolate the contribution of aspect-based retrieval, we compare \textsc{AspectSim} with the LBS baseline. The results demonstrate that, across all models, this setting yields substantially lower correlations (typically $<0.35$), with \textsc{DeepSeek-R1-32B} as a notable exception, which attains moderate correlation (0.495), likely due to its stronger multi-step reasoning capabilities. Nevertheless, even this model falls well below the performance of the proposed \textsc{AspectSim} ($\approx$0.58–0.59), underscoring the importance of explicitly extracting the aspect-relevant content.


\smallskip
\noindent\textbf{Can embeddings alone capture aspect-based similarity?}
As shown in Table~\ref{tab:baselines}, embedding-only baselines (WDS and PSD) exhibit weak correlations across all embeddings ($\approx$0.14--0.24), indicating that embeddings alone—without explicit aspect-based extraction—are insufficient for modeling aspect-based document similarity.



\begin{table}[!htb]
\centering
\small
\begin{adjustbox}{width=\linewidth,center}
  \begin{tabular}{c|cc||c|cc} 
    \toprule
   Embedding  & \textit{WDS} & \textit{PSD} & Embedding  & \textit{WDS} & \textit{PSD}\\
    \midrule
    MPNet & 0.207 & 0.162 & L-Mistral & \textbf{0.240} & 0.142  \\
    Jina & 0.220 & 0.174  & E5 & 0.229 & 0.146  \\
    Bilingual & 0.226 & 0.155  & Qwen3 & 0.229 & 0.155  \\
    Mxbai & 0.223 & 0.164  & Gemma & 0.223 & 0.160   \\
    S-Mistral & 0.228 & 0.141 \\
    \bottomrule
  \end{tabular}
\end{adjustbox}
\vspace{1mm}
\caption{Spearman Correlation of the Embedding-based baseline similarity methods. The highest score among these two methods is highlighted in bold.}
\label{tab:baselines}
\end{table}

\noindent\textbf{Overall,} retrieval-based approaches are the most effective for aspect-based similarity, particularly when paired with larger LLMs (27--72B) and strong embeddings, e.g., \textsc{E5}, \textsc{Qwen3}, and \textsc{Gemma}.

\subsubsection{Input Sensitivity Analysis}

To better understand how \textsc{AspectSim} behaves under varying input conditions, we analyze its performance across datasets, aspect lengths, and document-length combinations for the \textit{Retrieve-then-Embed} method, our best performer so far. 

\begin{figure}[!htb]
    \centering
    \includegraphics[width=0.48\textwidth]{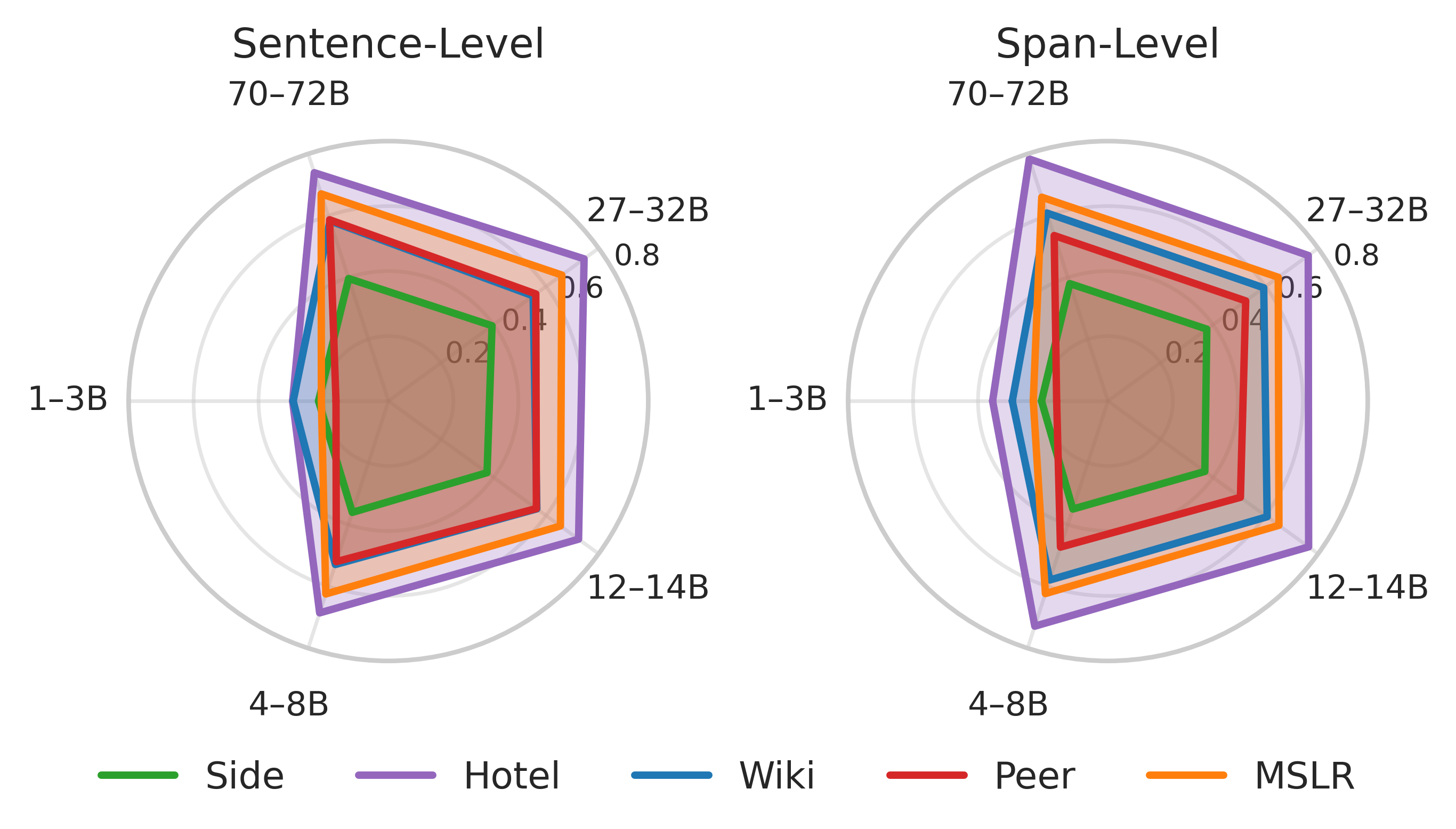}
    \vspace{-6mm}
    \caption{\textsc{AspectSim} performance across five datasets under Sentence-level and Span-level settings for different model-size ranges. Here, only the highest-performing embedding per dataset is shown.}
     \vspace{-3mm}
    \label{fig:dataset_perf}
\end{figure}

\noindent
\textbf{Dataset Trends.}
Figure~\ref{fig:dataset_perf} shows that correlations generally increase with model scale across datasets. \textsc{Hotel}, \textsc{MSLR}, \textsc{Wiki}, and \textsc{Peer} achieve stable performance (around 0.50), reflecting their well-structured, aspect-explicit content, whereas \textsc{Side} yields consistently lower correlations (0.30--0.40), highlighting the challenge of modeling ideologically diverse, loosely framed narratives.

\begin{figure}[!htb]
    \centering
    \includegraphics[width=0.46\textwidth]{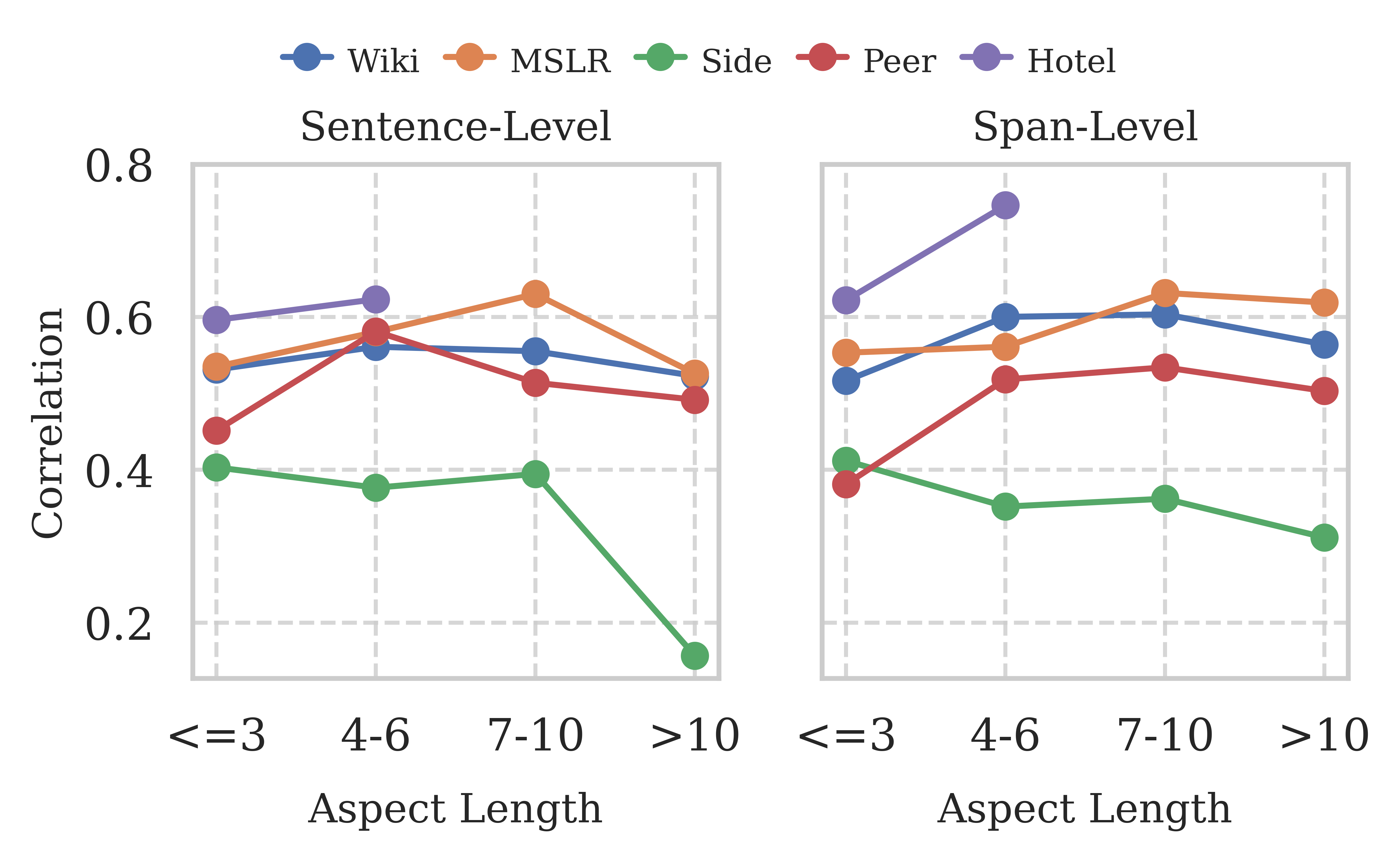}
     \vspace{-2mm}
    \caption{\textsc{AspectSim} performance across aspect lengths grouped by token count.}
    \vspace{-3mm}
    \label{fig:asplen_perf}
\end{figure}

\noindent
\textbf{Influence of Aspect Length.}
Figure~\ref{fig:asplen_perf} shows that aspect length has a mild effect on \textsc{AspectSim}: sentence-level performance peaks at moderate lengths ($\sim$10 tokens) and declines slightly thereafter, while span-level retrieval remains similarly stable. Comparable trends across document lengths indicate limited sensitivity to both aspect and document length (Appendix Figure~\ref{fig:doclen_perf}).

\begin{figure}
    \includegraphics[width=0.45\textwidth]{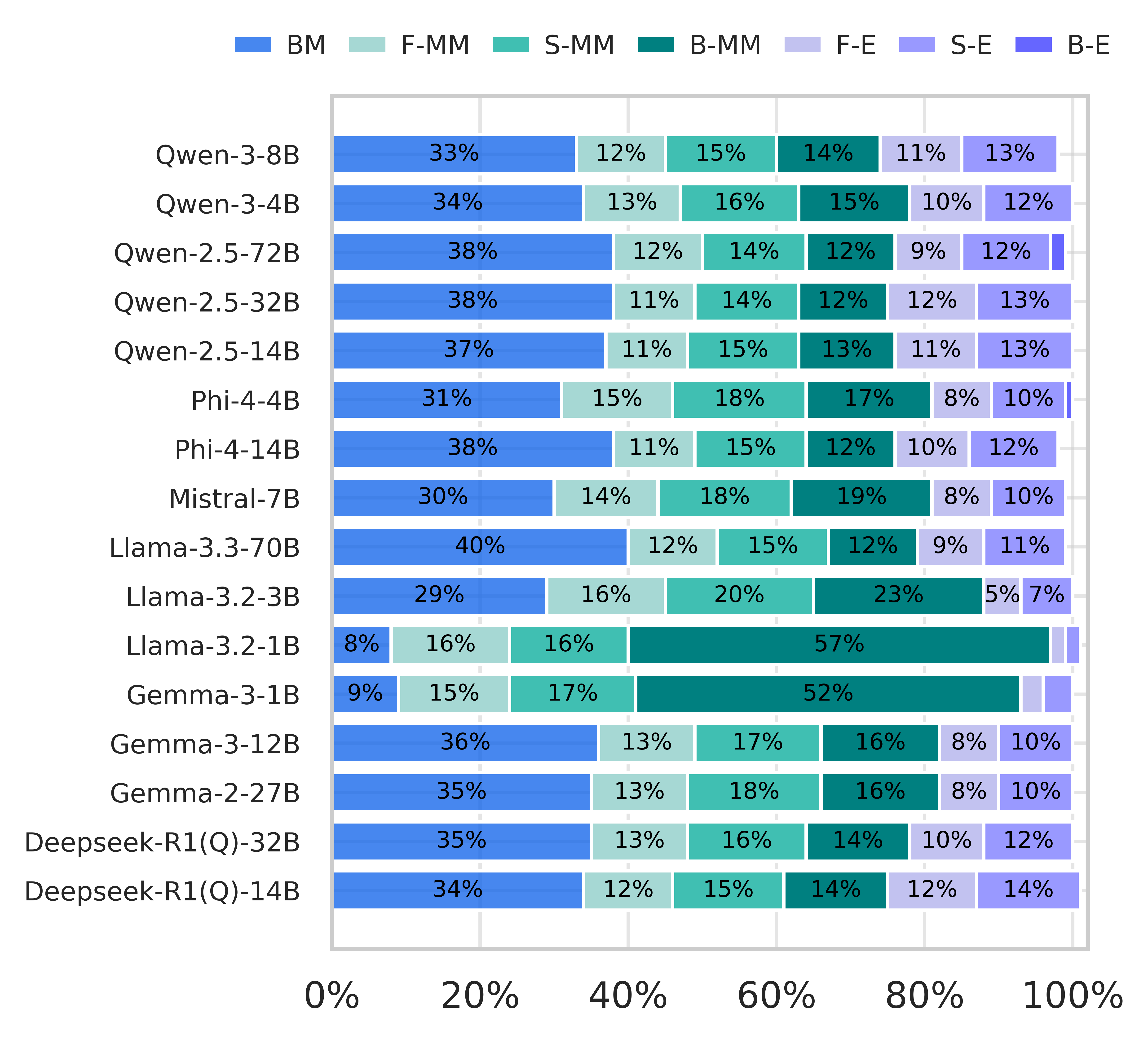}
\captionsetup{margin=0cm}
\vspace{-3mm}
  \caption{\textsc{AspectSim} retrieval accuracy across various LLMs at the sentence-level. BM denotes correctly retrieved aspect-relevant units. F-MM, S-MM, and B-MM indicate mismatched extractions for the first, second, and both documents, respectively. F-E, S-E, and B-E represent missed extractions (empty outputs) in the corresponding documents.}
\label{fig:retrieval_all} 
\vspace{-3mm}
\end{figure}

\begin{figure}[!htb]
       \centering
        \begin{subfigure}{0.4\textwidth}
            \includegraphics[width = \linewidth]{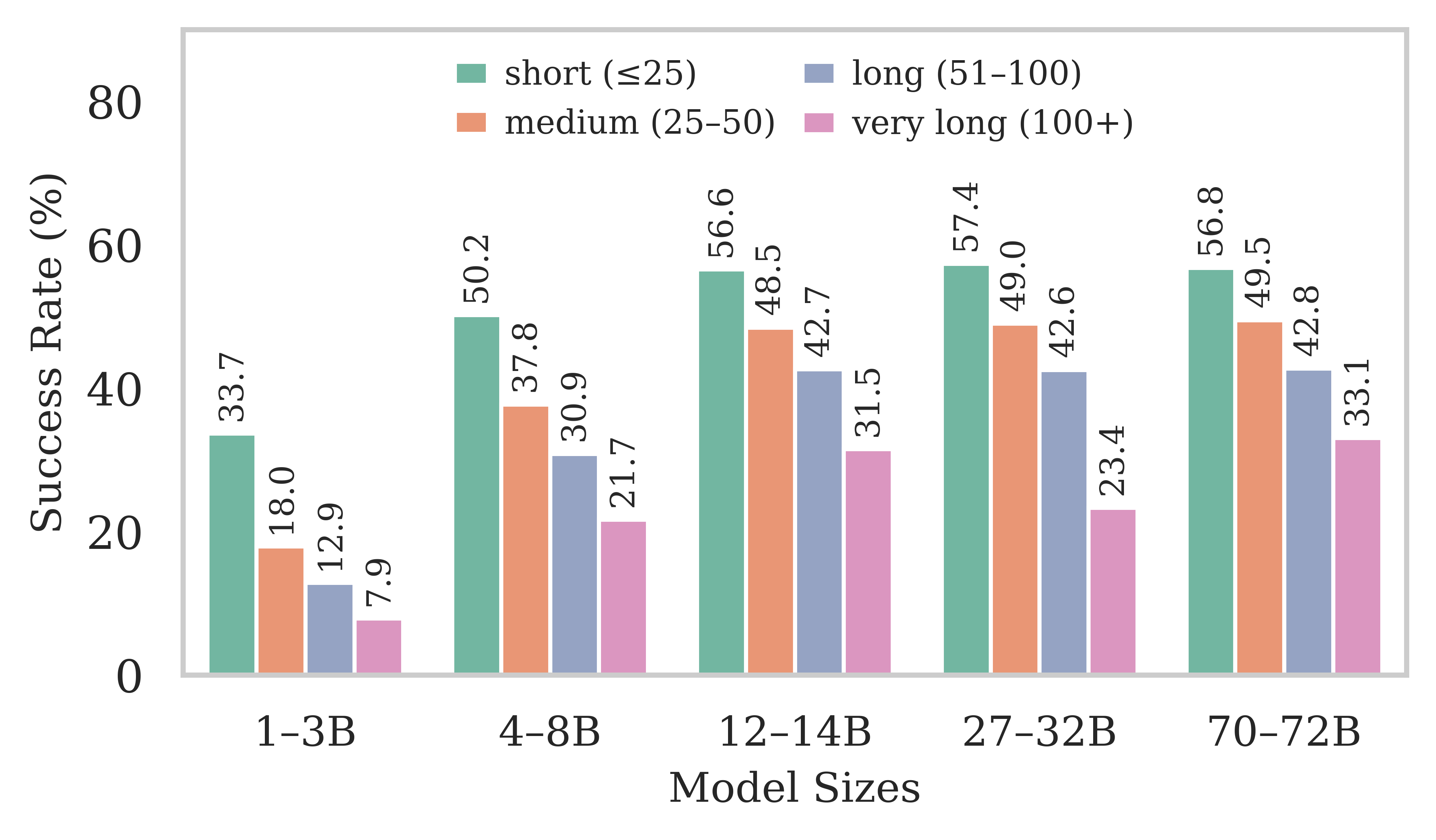}
            \centering            \captionsetup{justification=centering,margin=0cm}
            \vspace{-5mm}
             \caption{Document Size Effect}
            \label{subfig:retrieval_success_docs}            
        \end{subfigure}\hspace{0\textwidth}
         \begin{subfigure}{0.4\textwidth}
            \includegraphics[width = \linewidth]{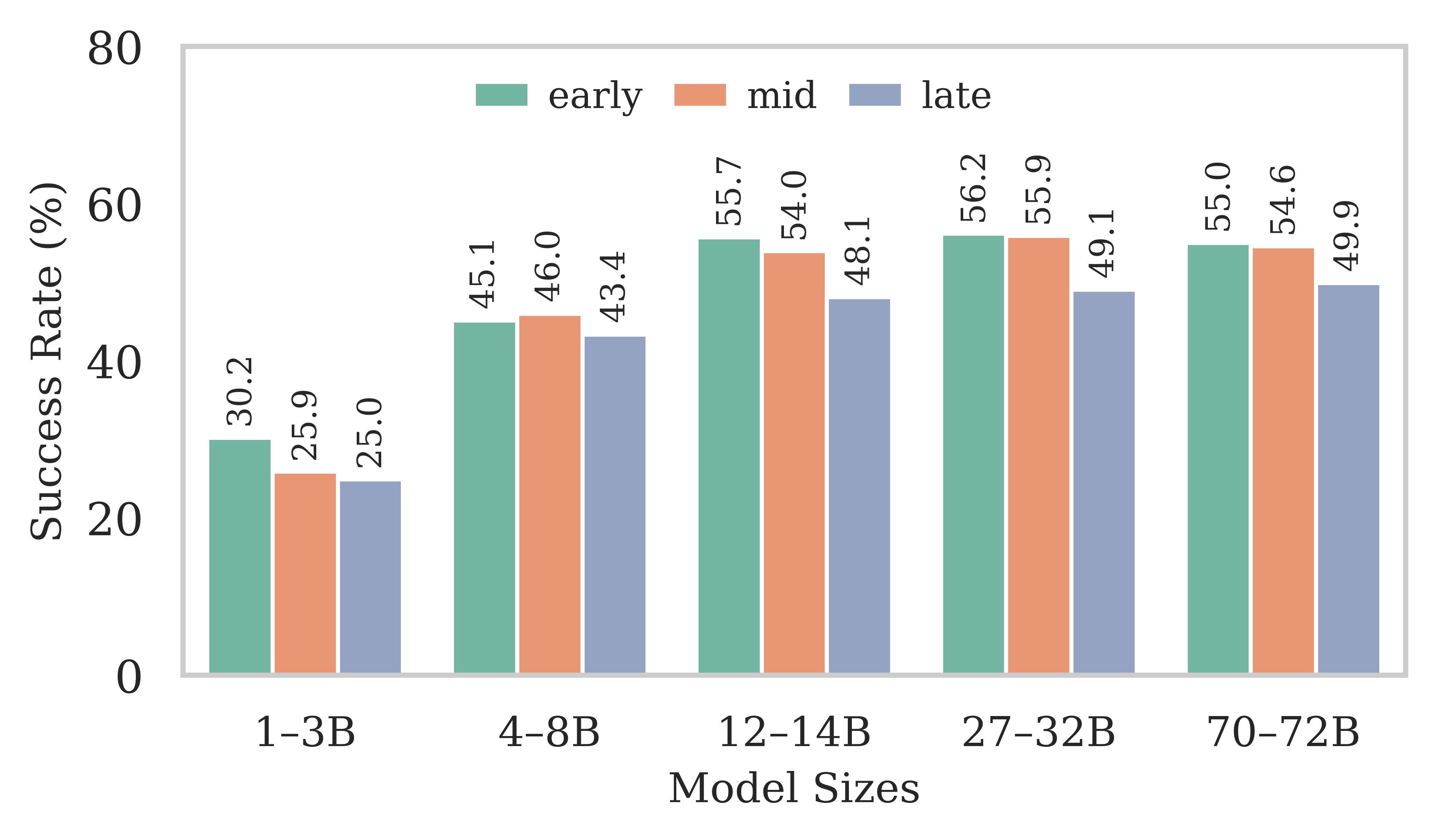}
            \centering            \captionsetup{justification=centering,margin=0cm}
            \vspace{-5mm}
            \caption{Sentence Position Effect}
            \label{subfig:retrieval_success_pos}            
        \end{subfigure}
        ~
        \vspace{-1mm}
        \caption{Effect of document length and sentence position on sentence-level retrieval success across model sizes.}
        \label{fig:retrieval_success} 
        \vspace{-3mm}
\end{figure}


\subsubsection{Retrieval Performance Analysis}
 
\smallskip
\noindent\textbf{How effectively do LLMs retrieve aspect-relevant evidence?} Figure~\ref{fig:retrieval_all} reports the retrieval accuracy of the \textit{Retrieve-Then-Embed} method under sentence-level settings. It is observed that the mid- and large-scale models  (e.g., Qwen-2.5-32B and LLaMA-3.3-70B) achieve the highest retrieval accuracy (approximately 38–40\%), while smaller models (i.e., Gemma-3-1B and LLaMA-3.2-1B) frequently miss or misidentify aspect-relevant sentences. Span-level retrieval shows similar results (Appendix Figure~\ref{subfig:retrieval-span}).




\smallskip

\noindent\textbf{How do document length and sentence position affect aspect-conditioned retrieval?} Figure~\ref{subfig:retrieval_success_docs} shows that sentence-level retrieval accuracy declines with increasing document length: short documents ($\leq$25 sentences) achieve the highest success rates, while performance drops sharply beyond 100 sentences, reflecting growing difficulty in retrieval from long context. Figure~\ref{subfig:retrieval_success_pos} further shows positional sensitivity, with smaller models struggling at mid and late positions and larger models exhibiting more uniform performance. Overall, document length and positional depth substantially affect retrieval precision, with model scaling only partially mitigating these challenges (see Appendix~\ref{appendix:retrieval_analysis} for details).


\smallskip
\noindent\textbf{How robust are LLMs in refraining from retrieval when no aspect-relevant content exists?}
Table~\ref{tab:robustness} reports performance in \textit{Not-Found} scenarios. Larger models (27--72B) achieve substantially higher robustness ($\approx 75-76\%$), reflecting a stronger ability to avoid false positives when no aspect-relevant evidence exists. Across datasets, \textsc{Hotel} and \textsc{MSLR} show the highest robustness (79.89\% and 77.74\%), while \textsc{Side} (50.94\%) and \textsc{Peer} (57.66\%) remain challenging due to their multi-perspective and argumentative content. Overall, smaller models tend to over-extract even in the absence of relevant evidence, whereas larger models are better at recognizing when to abstain, underscoring the role of scaling in improving precision in aspect-conditioned retrieval.
 

\begin{table}[!htb]
\vspace{-2mm}
\centering
\scriptsize
\begin{adjustbox}{width=0.9\linewidth,center}
\begin{tabular}{p{1.2cm}||ccccc>{\columncolor{gray!15}}c} 
    \toprule
    Model Size & Wiki & MSLR & Side & Peer & Hotel & Overall \\
    \midrule
    1-3B & 19.46 & 22.84 & 24.32 & 17.71 & 31.74 & 23.21\\
    4-8B & 64.49 & 89.85 & 57.23 & 67.10 & 90.01 & 73.74\\
    12-14B & 58.05 & 87.16 & 52.70 & 53.16 & 87.16 & 67.25\\
    27-32B & 66.20 & 92.73 & 58.89 & 71.55 & 93.39 & \textbf{76.55}\\
    70-72B & 68.16 & 91.95 & 56.89 & 68.25 & 93.65 & \textbf{75.38}\\
    \midrule
    \rowcolor{gray!15}
    Overall & 56.31 & \textbf{77.74} & 50.94 & 57.66 & \textbf{79.89}  \\
    \bottomrule
\end{tabular}
\end{adjustbox}
\vspace{-2mm}
\caption{\textsc{AspectSim} robustness (\%) across LLM sizes and datasets, measured as the proportion of ``Not-Found" cases where models correctly refrained from extracting any evidence.}
\vspace{-4mm}
\label{tab:robustness}
\end{table}

\vspace{-2mm}
\section{Conclusion}\vspace{-2mm}
In this work, we studied the importance of explicating aspects in measuring document similarity and introduced \textsc{AspectSim}, a new metric that estimates aspect-conditioned document similarity instead of holistic document similarity. Our results show that simple GPT-4o-prompting yields a reliable implementation of \textsc{AspectSim} with $\mathbf{\approx 90\%}$ correlation with human judgments. This result demonstrates the importance of explicitly accounting for aspects in measuring document similarity; the use of GPT-4o poses challenges in terms of \textit{accessibility} and \textit{reproducibility}, which are essential for scientific purposes. To make \textsc{AspectSim} accessible and reproducible, we leveraged relatively smaller open-source LLMs by directly prompting them, which turned out to be ineffective initially (\textbf{20–30\%} correlation). However, a simple refinement of \textsc{AspectSim} computation into a two-step process: \textit{extract} and \textit{embed}, demonstrated \textbf{140\%} higher correlation with human judgments (\textbf{58–59\%}). However, these results remain notably below the performance achieved by GPT-4–based models, suggesting that smaller open-source LLMs still lag behind very large proprietary models in capturing aspect-conditioned semantics.

\section*{Limitations}
The limitations of our work can be divided into two main categories: shortcomings of our aspect-conditioned dataset and limitations of the methods used for the similarity metric.

On the dataset side, although our benchmark spans five domains and includes around 26K aspect-conditioned document pairs, its overall scale and diversity remain limited. Due to the API costs associated with GPT-4o, we did not expand the dataset further. However, incorporating other proprietary systems, such as Gemini or Claude, into a multi-agent setup could yield a more varied set of examples, thereby enhancing diversity. Additionally, the aspects were generated from existing multi-document corpora, which may not fully reflect the range of user-defined or fine-grained perspectives encountered in practice. While approximately 10\% of the data were manually validated, the remaining annotations may still contain noisy or ambiguous similarity labels, particularly for borderline cases between `Somewhat' and `Marginally Similar'. Finally, because the dataset is restricted to English, it does not yet support multilingual or cross-lingual evaluation, which we leave for future work.

From a methodological perspective, our current implementation primarily relies on prompt-based aspect extraction and embedding-based similarity computation, without exploring post-training (i.e., supervised fine-tuning, reward-based fine-tuning, etc.) or retrieval-augmented variants that might improve the alignment. Furthermore, we experimented with only a limited set of prompting strategies, which may limit our understanding of how alternative prompt designs affect the metric's overall behavior. 
Finally, we have not yet examined how aspect-conditioned similarity could benefit downstream tasks such as retrieval reranking, fact verification, or hallucination detection, which remain promising directions for future work.


\bibliography{References/custom}

\appendix

\section*{Appendix}
\section{Additional Details}
\subsection{Data Selection Criteria}
\label{sec:sampling}

To ensure balanced coverage, we sampled approximately 2,100 document pairs across five domains. The \textsc{Peer} and \textsc{Hotel} datasets were included in full due to their moderate size and consistent pairwise similarity (0.6–0.9). For \textsc{Wiki}, \textsc{Side}, and \textsc{MSLR}, we selected samples from their respective test sets within the same similarity range to retain topical coherence while avoiding trivial or unrelated pairs. 


We limited the sample size to maintain \textbf{semantic relevance} and \textbf{computational feasibility} for the multi-stage annotation process. Aspect-conditioned labeling involves several steps that require GPT-4o calls over large contexts. Expanding beyond the current scale would substantially increase API cost and verification effort without yielding proportional gains in diversity or representativeness. 

Finally, we prioritized representational diversity over sheer volume. We sampled data across news, scientific, medical, political, and opinion domains, enabling a systematic evaluation of cross-domain generalization in open-form aspect-based document similarity. This diversification allows us to analyze whether design metrics exhibit consistent aspect-conditioned behavior across domains rather than within a single domain type.

\subsection{Dataset Statistics}\label{appen:stat}

\begin{table}[!b]
\centering

\begin{adjustbox}{max width=1\linewidth}
\begin{tabular}{lccccc}
\toprule
\textbf{Statistic} & \textbf{Side} & \textbf{Wiki} & \textbf{MSLR} & \textbf{Peer} & \textbf{Hotel} \\
\midrule
\multicolumn{6}{l}{\textbf{Label Stats}} \\[1pt]
\makecell[l]{Highly Similar}  & 3,551 & 3,382 & 875 & 592 & 65 \\
\makecell[l]{Somewhat Similar} & 3,041 & 2,551 & 3,133 & 532 & 211 \\
\makecell[l]{Marginally Similar} & 965 & 646& 1660 & 178 & 153 \\
\makecell[l]{Not Found} & 1300 & 1239 & 1306 & 360 & 96 \\
\midrule
\multicolumn{6}{l}{\textbf{Aspect Stats}} \\[1pt]
\makecell[l]{\# Aspects (avg. 1 sentence)} & 4,862 & 4,569 & 3,727 & 883 & 322 \\
\makecell[l]{\# Aspects (avg. $>$1 sentences)} & 3,995 & 3,249 & 3,247 & 779 & 203 \\
\rowcolor{gray!15}
Total No. of Aspects & 8,857 & 7,818 & 6,974 & 1,662 & 525 \\
\midrule
\multicolumn{6}{l}{\textbf{Doc Stats}} \\[1pt]
No. of Documents & 650 & 650 & 650 & 120 & 50\\
Avg. Doc 1 length & 27 & 23 & 12 & 20 & 8 \\
Avg. Doc 2 length & 43  & 18 & 11 & 20 & 7 \\
Avg. \# aspects/doc pair & 13  & 12 & 10 & 13 & 10 \\
Min Doc length & 11 & 11 & 11 & 11 & 11 \\
Max Doc length & 279 & 194 & 67 & 62 & 11 \\
\bottomrule
\end{tabular}
\end{adjustbox}
\caption{Dataset statistics comprising approximately \textbf{26K} aspects. \textit{Label Stats} present the distribution of similarity labels, while \textit{Aspect Stats} and \textit{Doc Stats} summarize average aspect-level and document-level measures across datasets. Here, Aspects (avg. $1$ sentence) correspond to cases where the relevant content in both documents is captured within a single sentence, while Aspects (avg. >1 sentence) represent multi-sentence spans.}
\label{tab:dataset_stats}
\end{table}

Table~\ref{tab:dataset_stats} summarizes the key statistics of our constructed dataset, which comprises approximately 26K aspect instances generated from 2,120 document pairs across five domains.  Overall, the label distribution remains relatively balanced, which provides adequate representation of both semantically aligned and divergent aspect pairs. On average, each document pair contains 10–13 aspects, with larger domains, such as \textsc{Side}, \textsc{Wiki}, and \textsc{MSLR}, contributing more than 6,000 aspects each. These aspects range from single-sentence mentions to multi-sentence, capturing the natural variation in how topics are discussed across documents. Furthermore, document lengths exhibit notable diversity, ranging from short news articles in \textsc{Wiki} (about 18–23 sentences) to longer argumentative or scientific texts in \textsc{Side} and \textsc{Peer} (20–40 sentences on average). This diversity in aspect density and document structure provides a broad and realistic testbed for evaluating aspect-conditioned document similarity across domains.

\subsection{The \textsc{AspectSim} Metric}\label{sec:aspetsim}
Given two documents, $D_A$ and $D_B$, and an aspect $a_i$, our goal is to quantify how similarly the documents discuss that aspect in a way that is aligned with human judgements. We propose \textsc{AspectSim}, a metric that operates in two stages: (1) Aspect-Conditioned Retrieval via LLM, and (2) Semantic Alignment via Embedding Similarity.

\subsubsection{Aspect-Based Retrieval with LLMs}
Directly prompting an LLM to “extract sentences relevant to the aspect” can often yield outputs that are loosely related, as the model may infer relevance even when the aspect is not discussed in the document. To mitigate this, we adopt a two-step prompting strategy that first verifies the presence of an aspect and then extracts the corresponding evidence, ensuring faithful and grounded retrieval.

\paragraph{Aspect Presence Verification.}
For each document $D \in {D_A, D_B}$, we first reformulate the aspect $a_i$ into a binary query of the form:

\textit{$Q_i$: “Does the document discuss $a_i$?”}

\noindent
The LLM then provides a yes/no response, indicating whether the aspect is discussed in the document. This verification step ensures that the model grounds its subsequent retrieval in actual document content, thereby preventing the extraction of unsupported or spurious evidence.

\paragraph{Evidence Extraction}
If the answer in the previous step is affirmative, then LLM is instructed to extract the most relevant segment $S_D^{(i)} \subseteq D$ that discusses $a_i$. We consider three design variations: retrieving the single most representative sentence, retrieving multiple relevant sentences (span-level), or summarizing the aspect-relevant content before similarity estimation. To maintain faithfulness and avoid cross-contamination, aspect-conditioned retrieval is performed independently for $D_A$ and $D_B$. Presenting both documents together could cause the LLM to mix evidence, extracting sentences from one document while processing the other. Therefore, independent processing ensures that each retrieval strictly reflects what is stated within its own document. This stage yields a pair of aspect-relevant evidence $(S_A^{(i)}, S_B^{(i)})$ that are directly interpretable and traceable to their source context. The prompts for design variations are provided in Appendix~\ref{appendix:design_prompts}.

\subsubsection{Embedding-Based Semantic Alignment}
To quantify semantic similarity between the retrieved evidence, we employ a sentence-level embedding model $\mathcal{E}(\cdot)$ to obtain vector representations of the extracted aspect-relevant text evidence:

\[
v_A^{(i)} = \mathcal{E}(S_A^{(i)}), \quad 
v_B^{(i)} = \mathcal{E}(S_B^{(i)}).
\]

The aspect-conditioned similarity score is then computed using cosine similarity:

\[
\textsc{AspectSim}(D_A, D_B, a_i) = 
\frac{v_A^{(i)} \cdot v_B^{(i)}}{\|v_A^{(i)}\| \, \|v_B^{(i)}\|}.
\]
This embedding-based alignment provides a direct, interpretable method for measuring how two documents discuss a specific aspect. 

Together, the two stages allow \textsc{AspectSim} to move beyond opaque LLM judgments and produce similarity scores that are transparent and grounded in textual evidence.

\subsection{Implementation Setup and Model Identifiers}
All the experimented models were obtained from the HuggingFace library. For the LLMs, we used the \texttt{instruct} variants of each model and performed inference with vLLM. During inference, we use each model’s default decoding configuration, including temperature, sampling strategy, and other hyperparameters. For the embeddings, we used the following Hugging Face \textit{Sentence-Transformers} model identifiers: 
\begin{itemize}[leftmargin=*, itemsep=1pt, topsep=1pt]
\item \texttt{sentence-transformers/all-mpnet-base-v2}
\item \texttt{jinaai/jina-embeddings-v3}
\item \texttt{Lajavaness/bilingual-embedding-large}
\item \texttt{mixedbread-ai/mxbai-embed-large-v1}
\item \texttt{Salesforce/SFR-Embedding-Mistral}
\item \texttt{Linq-AI-Research/Linq-Embed-Mistral}
\item \texttt{intfloat/e5-large}
\item \texttt{Qwen/Qwen3-Embedding-8B}
\item \texttt{google/embeddinggemma-300m}

\end{itemize}

\subsection{Projection Similarity Difference Method}
\label{appendix:psd}
Given two document embeddings $\mathbf{d}_1, \mathbf{d}_2 \in \mathbb{R}^n$ and an aspect embedding $\mathbf{a} \in \mathbb{R}^n$, we first compute each document’s projection onto the aspect vector as:
\begin{equation}
\text{proj}(\mathbf{d}_i, \mathbf{a}) = \frac{\mathbf{d}_i \cdot \mathbf{a}}{\|\mathbf{a}\|}
\end{equation}
where $\mathbf{d}_i \cdot \mathbf{a}$ denotes the dot product and $\|\mathbf{a}\|$ is the L2 norm of the aspect embedding.

We then define the \emph{projection difference} between the two documents as:
\begin{equation}
\Delta(\mathbf{a}, \mathbf{d}_1, \mathbf{d}_2) = \biggl|\, \frac{\mathbf{d}_1 \cdot \mathbf{a}}{\|\mathbf{a}\|} - \frac{\mathbf{d}_2 \cdot \mathbf{a}}{\|\mathbf{a}\|} \biggr|.
\end{equation}
This absolute difference captures the magnitude of disagreement between the documents with respect to the given aspect, independent of which document is more strongly aligned.  A smaller value of $\Delta(\mathbf{a}, \mathbf{d}_1, \mathbf{d}_2)$ indicates higher aspect-specific alignment, whereas a larger value denotes lower alignment.

Since this measure is inversely related to similarity, we normalize and invert it to obtain a positively correlated similarity score:
\begin{equation}
\texttt{Sim}(\mathbf{a}, \mathbf{d}_1, \mathbf{d}_2) = 1 - \frac{\Delta(\mathbf{a}, \mathbf{d}_1, \mathbf{d}_2)}{Z}
\end{equation}

\begin{equation}
Z = \max_{(\mathbf{a}, \mathbf{d}_1, \mathbf{d}_2) \in \mathcal{D}} \Delta(\mathbf{a}, \mathbf{d}_1, \mathbf{d}_2),
\end{equation}

where $Z$ denotes a normalization constant, computed as the maximum projection difference observed across all document pairs in the dataset. This ensures that $\texttt{Sim}(\mathbf{a}, \mathbf{d}_1, \mathbf{d}_2) \in [0, 1]$, where higher values denote greater aspect-conditioned similarity.


\begin{figure}[!b]
    \centering
     \vspace{-4mm}
    \includegraphics[width=0.48\textwidth]{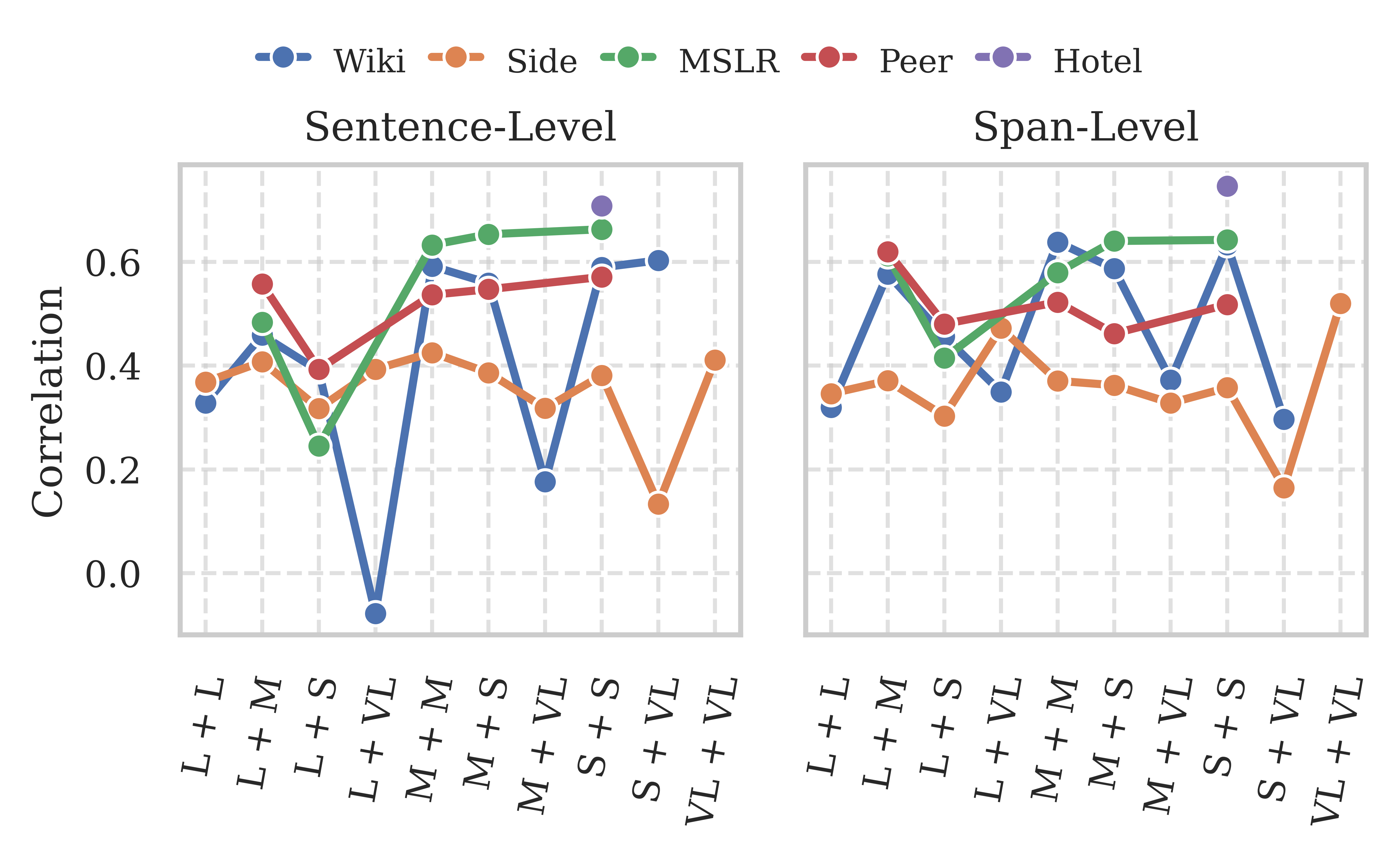}
    \caption{\textsc{AspectSim} performance across different document-size pairings. Abbreviations: S (short:$\leq25$), M (medium:25-50), L (long:51-100), VL (very long:$100+$).}
    \label{fig:doclen_perf}
\end{figure}

\begin{figure}
    \includegraphics[width=0.45\textwidth]{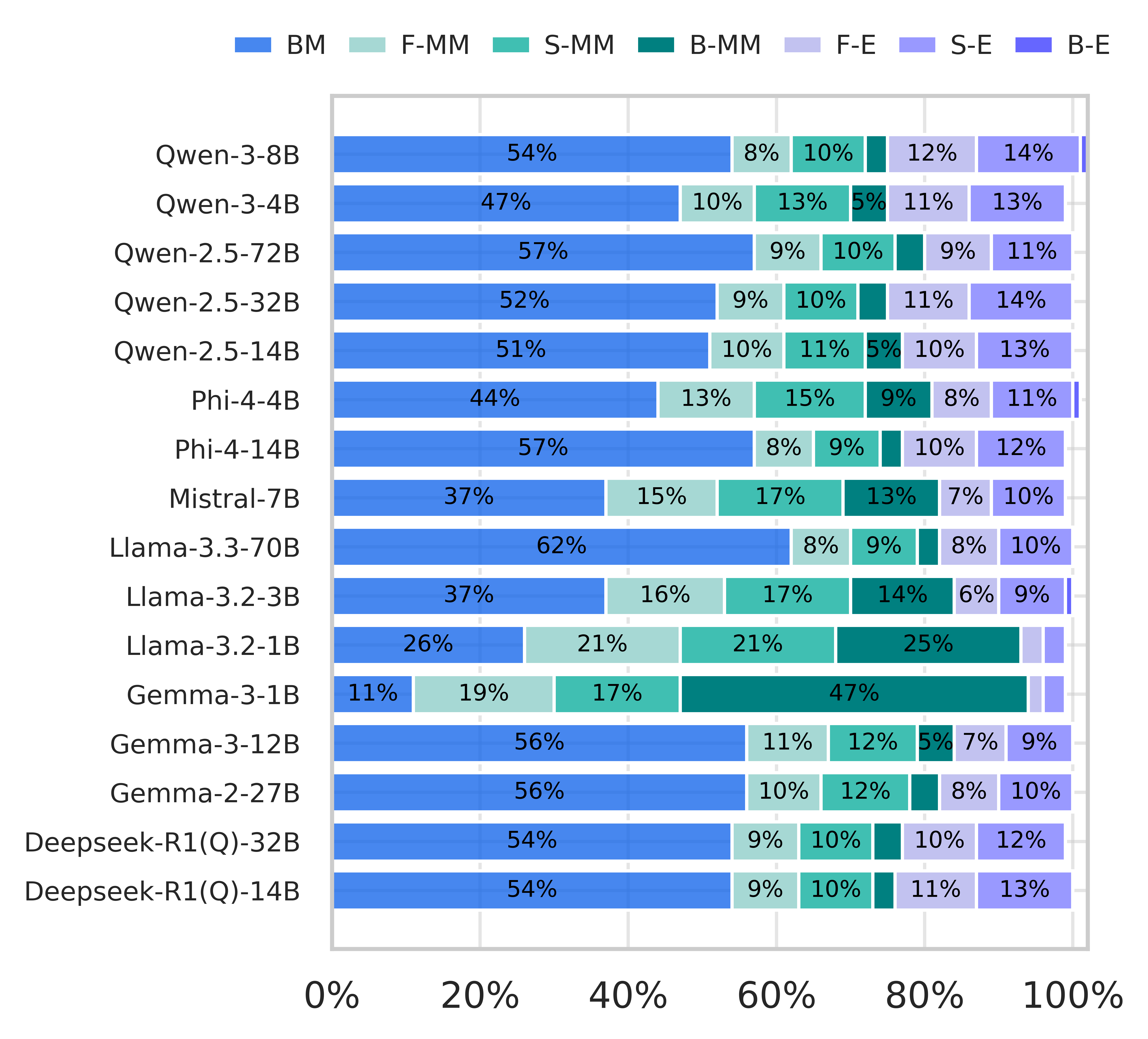}
\captionsetup{margin=0cm}
  \caption{\textsc{AspectSim} retrieval accuracy across various LLMs at the span-level. It shows substantially higher “BM” rates, often exceeding 50\%. This apparent improvement is largely due to multi-sentence extraction, which increases the likelihood of overlap when gold annotations are single sentences.}
\label{subfig:retrieval-span} 
\end{figure}

\begin{figure*}[!t]
        \begin{subfigure}{0.48\textwidth}
            \includegraphics[width = \textwidth]{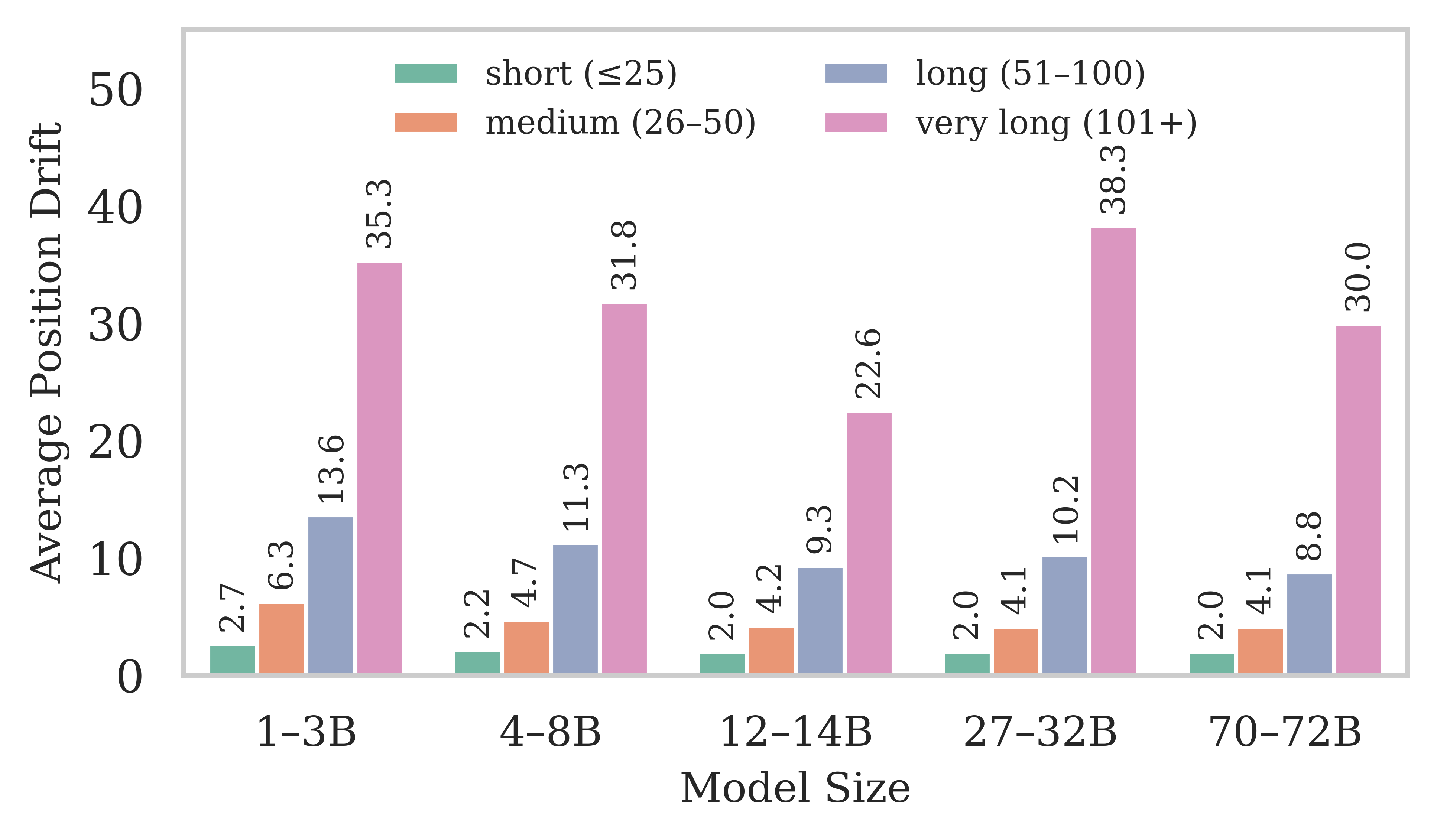}
            \centering            \captionsetup{justification=centering,margin=0cm}
            \caption{Error across different document sizes}
            \label{subfig:retrieval_error_docs}            
        \end{subfigure}\quad
        \begin{subfigure}{0.48\textwidth}
            \includegraphics[width = \textwidth]{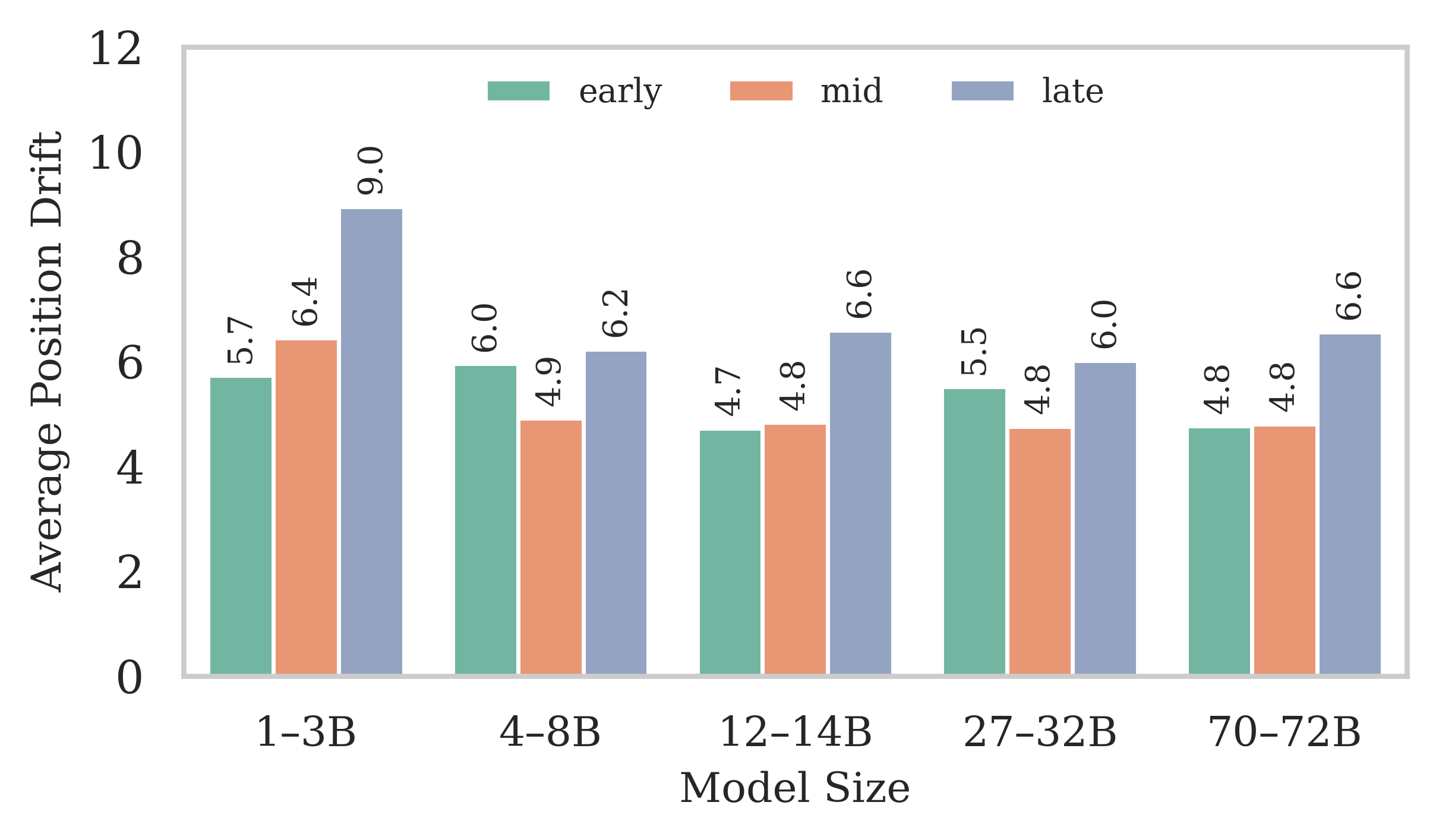}
            \centering            \captionsetup{justification=centering,margin=0cm}
             \caption{Error across different sentence positions}
            \label{subfig:retrieval_error_pos}            
        \end{subfigure}
        ~
        \caption{Average position drift between LLM-extracted and ground-truth aspect-relevant sentences across (a) different document sizes and (b) sentence positions in the document.}
        \label{fig:retrieval_error} 
\end{figure*}

\begin{figure*}[!t]
       
        \begin{subfigure}{0.48\textwidth}
            \includegraphics[width = \textwidth]{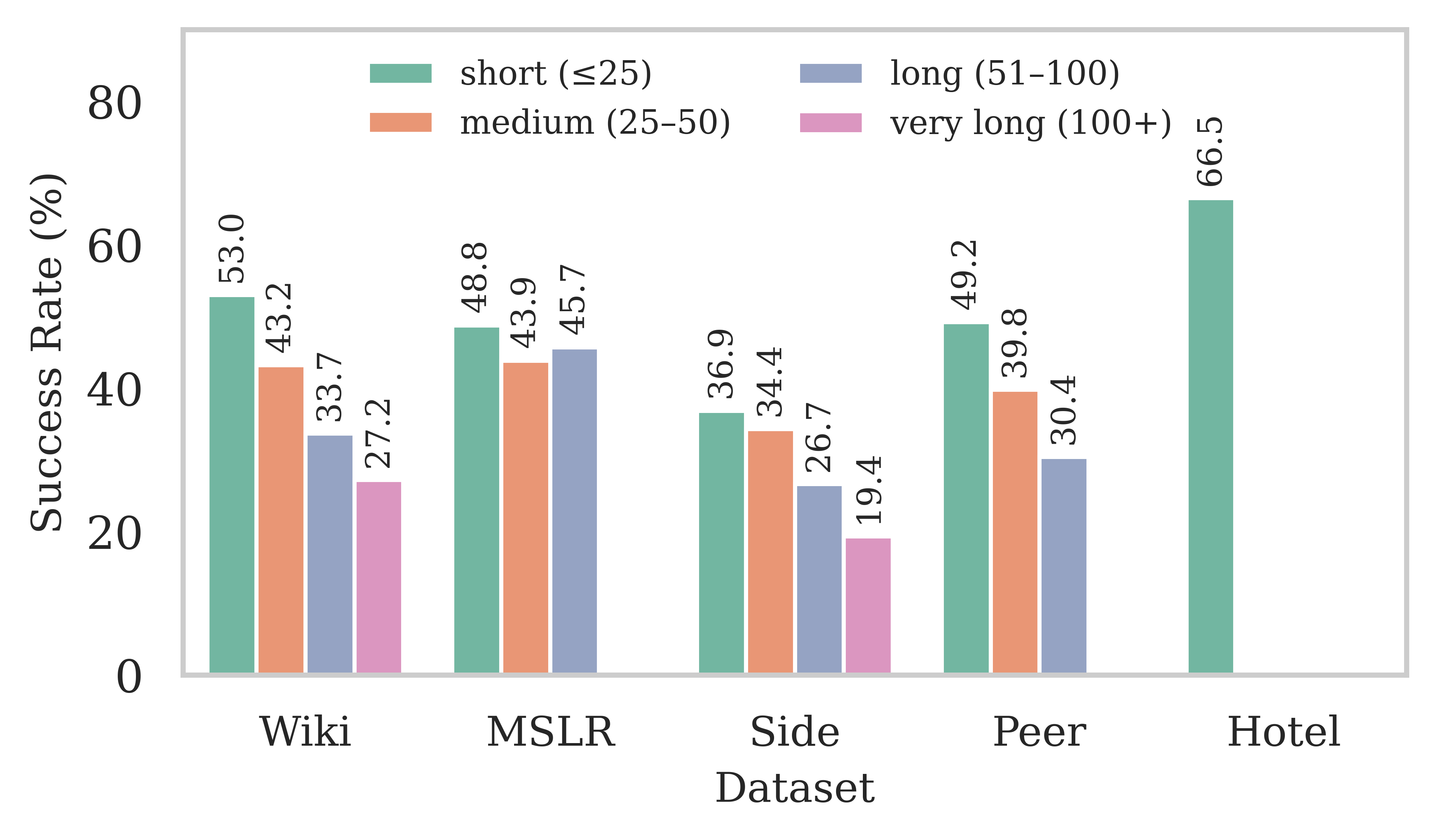}
            \centering            \captionsetup{justification=centering,margin=0cm}
             \caption{Document Size Effect}
            \label{subfig:retrieval_success_docs_dataset}            
        \end{subfigure}\quad
         \begin{subfigure}{0.48\textwidth}
            \includegraphics[width = \textwidth]{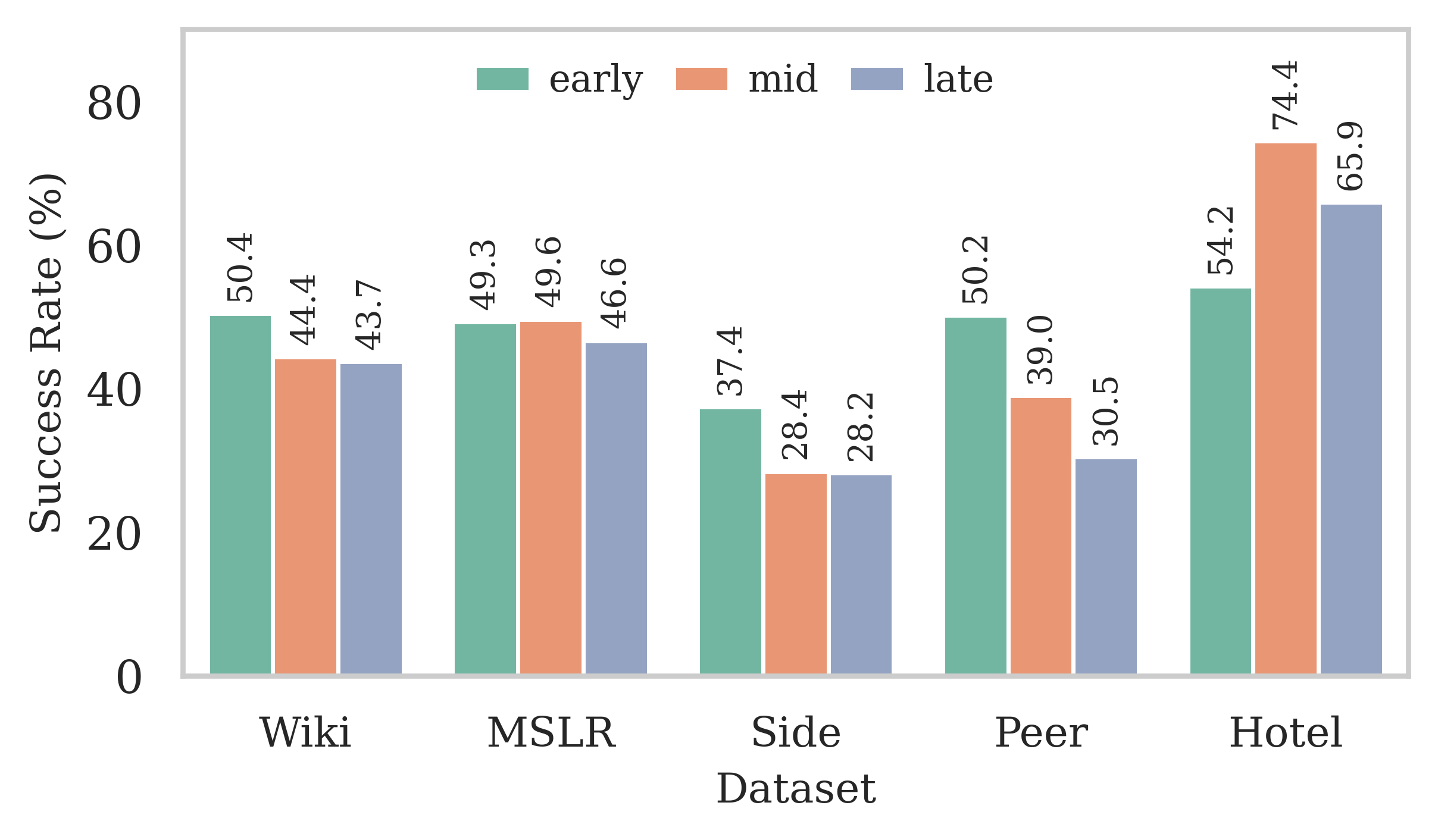}
            \centering            \captionsetup{justification=centering,margin=0cm}
            \caption{Sentence Position Effect}
            \label{subfig:retrieval_success_pos_dataset}            
        \end{subfigure}
        ~
        \caption{Effect of document length and sentence position on sentence-level retrieval success across datasets.}
        \label{fig:retrieval_success_dataset} 
\end{figure*}

\begin{figure*}[!t]
        \begin{subfigure}{0.48\textwidth}
            \includegraphics[width = \textwidth]{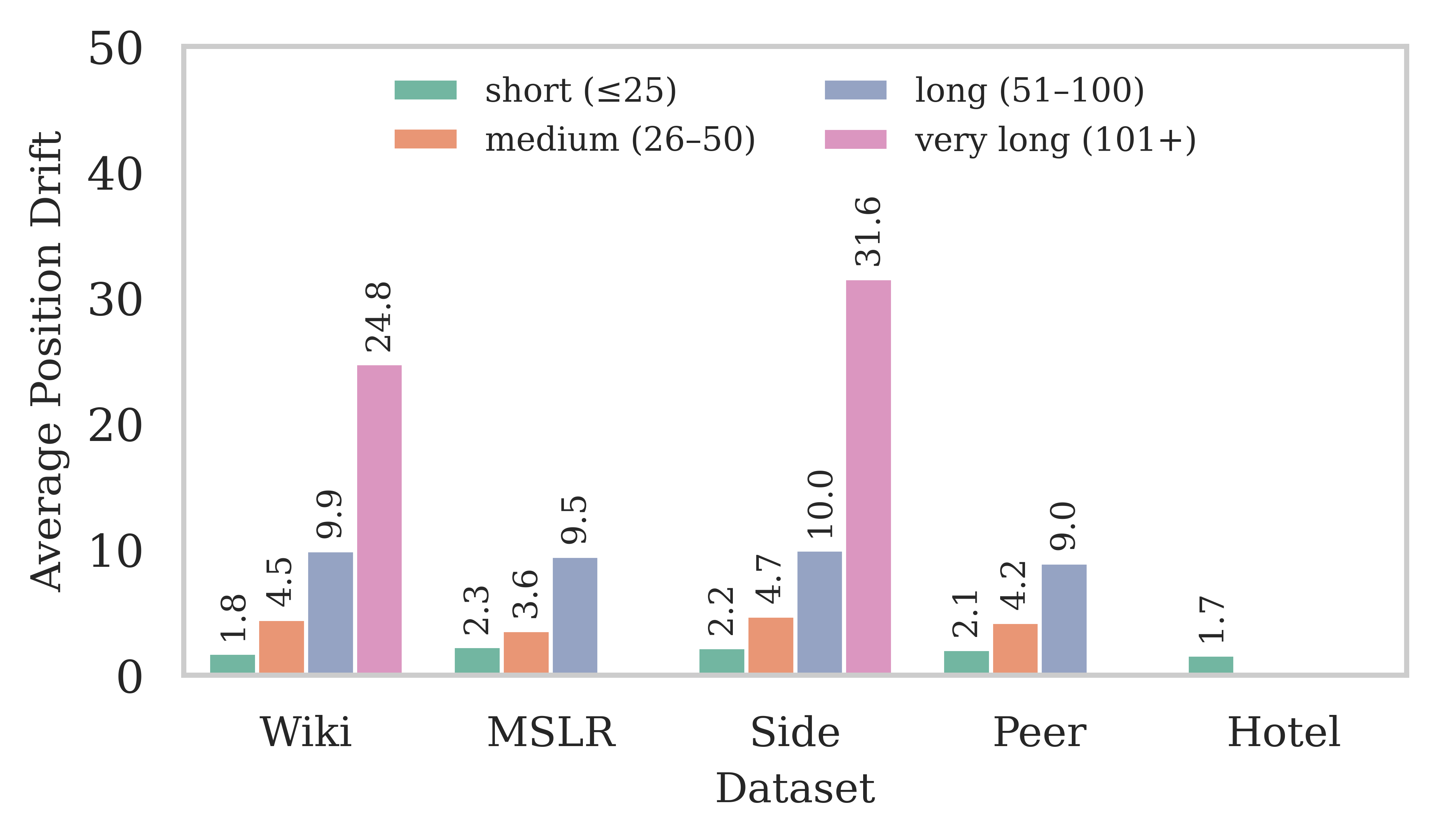}
            \centering            \captionsetup{justification=centering,margin=0cm}
            \caption{Error across different document sizes}
            \label{subfig:retrieval_error_docs_dataset}            
        \end{subfigure}\quad
        \begin{subfigure}{0.48\textwidth}
            \includegraphics[width = \textwidth]{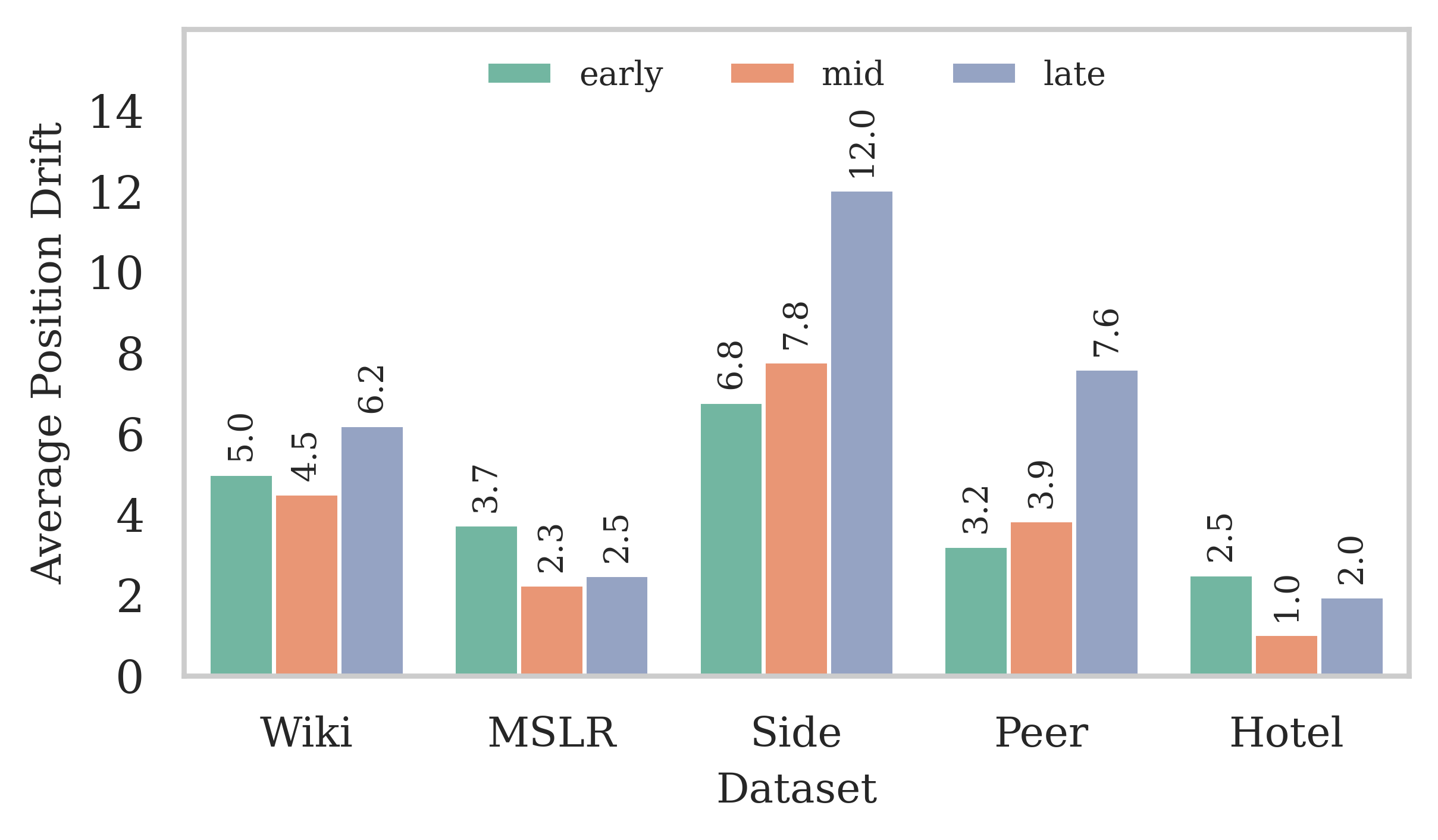}
            \centering            \captionsetup{justification=centering,margin=0cm}
             \caption{Error across different sentence positions}
            \label{subfig:retrieval_error_pos_dataset}            
        \end{subfigure}
        ~
        \caption{Average position drift between LLM-extracted and ground-truth aspect-relevant sentences for different datasets across (a) different document sizes and (b) sentence positions in the document.}
        \label{fig:retrieval_error_dataset} 
\end{figure*}

\subsection{Additional Results}
\label{sec:additional_result}

\subsubsection{Input Sensitivity of \textsc{AspectSim}}
\label{appendix:input_factors}
\noindent
\textbf{Impact of Document Lengths. }
Figure~\ref{fig:doclen_perf} illustrates the effect of document-length pairings on \textsc{AspectSim}. The \textsc{MSLR} and \textsc{Peer} datasets maintain consistently high correlations across most length combinations, whereas the \textsc{Wiki} dataset shows greater variability depending on document pairing. Across datasets, span-level retrieval yields slightly smoother trends and modest improvements ($\approx$ +0.02–0.04) over sentence-level retrieval. 


\subsubsection{Analysis on Retrieval Performance }
\label{appendix:retrieval_analysis}

\textbf{To what extent do LLMs deviate from the true positions of aspect-relevant sentences?}
Figure~\ref{fig:retrieval_error} quantifies positional error in retrieved sentences. As shown in Figure~\ref{subfig:retrieval_error_docs}, average position drift increases sharply with document length, reaching approximately 35–40 sentences for smaller models on very long documents. Larger models reduce this drift but still exhibit substantial error. Figure~\ref{subfig:retrieval_error_pos} shows that deviations are consistently higher for late-position sentences across all model sizes, highlighting persistent challenges LLMs faces in maintaining precise aspect grounding in long, late document contexts 


\textbf{How do dataset characteristics influence LLMs’ aspect-conditioned retrieval performance?}
Figures~\ref{fig:retrieval_success_dataset} and \ref{fig:retrieval_error_dataset} extend the earlier analysis presented in Figures~\ref{fig:retrieval_success} and \ref{fig:retrieval_error} by comparing retrieval behavior across datasets. We have observed that retrieval accuracy consistently decreases with longer documents (Figure~\ref{subfig:retrieval_success_docs_dataset}) and when the aspect-relevant sentence positions are in \textit{later} part of the documents (Figure~\ref{subfig:retrieval_success_pos_dataset}). This trend mirrors the model-scale findings in Figure~\ref{fig:retrieval_success}, reaffirming that increased context and \textit{late} positions introduce retrieval noise and distract the model from aspect-relevant content. Among all the datasets, \textsc{Hotel} achieves the highest accuracy, benefiting from short and structured inputs, while \textsc{Side} remains the most challenging due to its long, entangled narratives. \textsc{Wiki} and MSLR exhibit moderate, relatively stable performance, suggesting that well-organized informational content aids retrieval. Although performance on the \textsc{Peer} dataset is better than on the \textsc{Side} dataset, it still struggles with contextual dispersion, likely due to the argumentative nature of the texts. Figure~\ref{fig:retrieval_error_dataset} complements this by showing positional drift patterns. Longer documents, especially in the \textsc{Side} and \textsc{Wiki} datasets, cause substantial drift, whereas the errors remain stable across datasets for other document lengths. Furthermore, when the sentences are in the later part of the documents (Figure~\ref{subfig:retrieval_error_pos_dataset}), the error deviation increases, reflecting weaker reasoning in extended contexts when conditioned with aspects.

Overall, dataset characteristics such as narrative structure, narrative density, and document length significantly impact retrieval accuracy. These results highlight that \textsc{AspectSim}’s performance is shaped not only by model capacity but also by the inherent organization and discourse patterns of the dataset.

\textbf{Under what conditions do LLMs miss aspect-relevant content despite its presence?}
Table~\ref{tab:missed-retrieval} reports the overall missed retrieval rates aggregated across all models and datasets. Most missed cases occur in short documents (13.16\%), suggesting that limited context provides fewer cues for identifying aspect relevance. In contrast, the relatively lower missed rate in long documents indicates that models often extract some content, although not necessarily the correct one, which reflects a tendency to retrieve loosely related sentences (as supported by observations seen in Figures~\ref{subfig:retrieval_success_docs}, and \ref{subfig:retrieval_success_docs_dataset}) rather than making precise, aspect-grounded selections. Moreover, the error rate rises notably for late-position sentences (14.10\%), implying that as aspect-relevant content appears further down in the document, models either drift toward irrelevant regions or fail to locate the correct evidence altogether. This pattern aligns with the positional drift trends observed earlier (Figures~\ref {subfig:retrieval_success_pos} and \ref{subfig:retrieval_success_pos_dataset}), underscoring the challenge of maintaining focused retrieval over extended contexts.

\begin{table}[!htbp]
\centering
\small
\begin{tabular}{p{3cm}c}
    \toprule
    Document Size & \textit{Missed Rate}  \\
    \midrule
    Short & 13.16  \\
    Medium & 7.11  \\
    Long & 5.81  \\
    Verly Long & 7.77  \\
    \bottomrule
    \toprule
    Sentence Position & \textit{Missed Rate}  \\
    \midrule
    Early & 6.39  \\
    Mid & 9.29  \\
    Late & 14.10  \\

    \bottomrule
\end{tabular}

\vspace{1mm}
\caption{Percentage (\%) of missed retrieval across different document sizes and sentence positions calculated over all LLMs.}
\label{tab:missed-retrieval}
\end{table}

\section{Annotation Guidelines for Aspect-Based Document Similarity}
\label{appendix:annot-aspect}
\subsection{Annotation Task Overview}

Annotators are provided with: (i) A pair of documents (\textit{Document 1} and \textit{Document 2}) and (ii) A specific \textit{aspect}/\textit{rubric}/\textit{criteria} based on which the similarity of the two documents is evaluated.\\

\noindent
The task consists of the following steps:\\

\noindent
\textbf{Step 1:} Read the two documents carefully and understand the broad idea, interpretation, stance, and conclusions, etc.

\noindent
\textbf{Step 2:} Read the Aspect to understand what the aspect refers to and what kind of information could be relevant. 

\noindent
\textbf{Step 3:} Identify the most relevant sentence(s) from each document that relate to the given aspect. Compare the selected texts in terms of meaning, level of detail, and emphasis.

\noindent
\textbf{Step 4:} Assign a similarity label from one of the four predefined categories described below.

\noindent
\textbf{Step 5:} Provides a brief comment (optional) explaining the decision, especially for borderline cases.

\subsection{Similarity Labels and Definitions}
\paragraph{\underline{Highly Similar:}}
The selected texts from both documents express the same core idea, showing strong alignment in content, perspective, and emphasis, possibly using different wording or phrasing.

\noindent
\textbf{Indicators:}
\begin{itemize}
    \item Same claims, conclusions, or key facts
    \item No differences in stance or emphasis
    \item Paraphrased or reworded content
\end{itemize}

\noindent
\textbf{Example:}  
\textit{Aspect: Leadership Transparency}  
\begin{itemize}
    \item Doc 1: ``The CEO emphasized openness in all internal communications.''
    \item Doc 2: ``Leadership prioritized transparent updates to staff throughout the process.''
\end{itemize}

\noindent
\textbf{Comment:} Both texts describe leadership ensuring openness, conveying the same meaning using different phrasing.

\paragraph{\underline{Somewhat Similar:}} 
Both texts discuss the same general aspect and share substantial overlap, but differ in specific details, examples, or emphasis.

\noindent
\textbf{Indicators:}
\begin{itemize}
    \item Reference to the same aspect or topic with different levels of detail
    \item No contradiction, but incomplete alignment
\end{itemize}

\noindent
\textbf{Example:}  
\textit{Aspect: Project Delay}  
\begin{itemize}
    \item Doc 1: ``Supply chain issues led to a two-month delay.''
    \item Doc 2: ``The project timeline shifted slightly due to unexpected disruptions.''
\end{itemize}

\noindent
\textbf{Comment:} Both mention a delay, but only the first specifies the cause and duration.

\paragraph{\underline{Marginally Similar:}}
The texts are loosely related within the same broad topic but exhibit weak alignment or divergent perspectives.

\noindent
\textbf{Indicators:}
\begin{itemize}
    \item One text may focus on a different subpoint
    \item Conflicting interpretations, tone, or implications
\end{itemize}

\noindent
\textbf{Example:}  
\textit{Aspect: Team Collaboration}  
\begin{itemize}
    \item Doc 1: ``The team worked across departments to meet the deadline.''
    \item Doc 2: ``Lack of coordination led to multiple missed deadlines.''
\end{itemize}

\noindent
\textbf{Comment:} Both discuss team dynamics, but one is positive while the other is critical, reflecting conflicting outcomes.

\paragraph{\underline{Not Found:}} 
The aspect is discussed in only one document. The other document does not reference the aspect, either explicitly or implicitly.

\section{Annotation Guidelines for Document-Level Similarity (Aspect-Agnostic)}
\label{appendix:annot-aspect-free}

\subsection{Annotation Task overview}
Annotators are given \textit{Document 1} and \textit{Document 2}, and the goal is to assess how similar the two documents are in terms of their overall informational content. Annotators are instructed to follow the given steps :

\noindent
\textbf{Step 1:}  
Read both documents thoroughly to understand their overall content and intent.

\noindent
\textbf{Step 2:}  
For each document, identify the main topic and assertions, findings, outcomes, and conclusions.

\noindent
\textbf{Step 3:}  
Assess the extent to which the identified topics and information overlap between the two documents.

\noindent
\textbf{Step 4:}  
Select the most appropriate similarity label based on the definitions below.

\noindent
\textbf{Step 5: }  
Provide a brief explanation (optional) if the decision is borderline between two labels.

\subsection{Similarity Labels and Definitions}

\paragraph{\underline{Highly Similar:}} 
Both documents discuss the same primary topic and express the same central claims or conclusions, with significant overlap, either in the same wording or in paraphrased form. 


\paragraph{\underline{Somewhat Similar:}} 
Both documents discuss the same primary topic but overlap only partially in their core claims or supporting information. One document may include additional details, context, or emphasis not present in the other.

\paragraph{\underline{Marginally Similar:}}
The documents share only a broad theme/event, but differ in their main focus, claims, or conclusions. Any overlap is weak, indirect, or high-level.

\paragraph{\underline{Not Similar:}}
The documents do not share a primary topic, central claim, or conclusion, and there is no meaningful informational overlap.


\onecolumn
\section{Data Generation Prompts}
\label{appendix:prompts}




\scriptsize
\setlength{\tabcolsep}{0pt}
\renewcommand{\arraystretch}{1}

\begin{longtable}{p{\textwidth}}

\begin{tcolorbox}[
    colback=gray!5,
    colframe=blue!60!black,
    title=\textbf{Prompt Template for GPT-4o Based Aspect Similarity Data Generation},
    fonttitle=\bfseries,
    sharp corners,
    boxrule=0.8pt
]
{\ttfamily
 You are an intelligent assistant responsible for identifying shared aspects between two documents and extracting semantically related sentences or paragraphs corresponding to those aspects.\\

     \textbf{\#\#\#**Your Task:**}\\
      1. Identify Relevant Aspects or Topics:
      \begin{itemize}
      \renewcommand\labelitemi{-}
          \item Read both documents carefully.
          \item Identify key aspects or themes that are discussed in both documents and must have relevant sentences for those aspects or themes.
      \end{itemize}

      2. Extract Corresponding Sentence Pairs:
      \begin{itemize}
      \renewcommand\labelitemi{-}
          \item  For each identified aspect, extract corresponding sentences or paragraphs from **Document 1** and **Document 2** that discuss the same aspect.
          \item Each extracted pair **must come from different documents**.
      \end{itemize}
      
      3. Categorize Similarity:\\
        Compare each extracted pair and classify them into one of the following categories:
        \begin{itemize}
        \renewcommand\labelitemi{-}
            \item Highly Similar: The sentences focus on the same key points, with the same emphasis and content, though possibly using different wording or framing. The core ideas, details, and perspectives align closely.
            \item Somewhat Similar: The sentences discuss the same key aspects and points, with substantial overlap in meaning, but may differ in details or depth. There should be no conflicting aspects.
            \item Marginally Similar: The sentences discuss **different** or **conflicting** ideas. Even if they belong to the same broad topic, their focus or perspectives are misaligned.
        \end{itemize}

      4. Provide Explanation:
      \begin{itemize}
      \renewcommand\labelitemi{-}
          \item Justify why each pair was classified as **Highly Similar, Somewhat Similar, or Marginally Similar**.\\
      \end{itemize}

      \textbf{\#\#\# **Expected Output Format:**}
      Give at most 15 unique aspect pairs. Your response must be a **single valid JSON object** structured as follows:
\begin{verbatim}
```json
[
  {
    "aspect": "Identified aspect or criteria",
    "pairs": {
      "sentence1": "Sentence(s) from Document 1",
      "sentence2": "Sentence(s) from Document 2",
      "similarity": "Highly Similar | Somewhat Similar | Marginally Similar",
      "reason": "Brief explanation of the similarity classification"
  }}
  ......
]
```
\end{verbatim}

\begin{verbatim}
Input Documents:
    {Document 1}
    {Document 2}
\end{verbatim}

}
\end{tcolorbox}
\\

\begin{tcolorbox}[
    colback=gray!5,
    colframe=blue!60!black,
    title=\textbf{Prompt Template for GPT-4o Based Negative Instance Generation},
    fonttitle=\bfseries,
    sharp corners,
    boxrule=0.8pt
]

{\ttfamily
You are an intelligent extractor tasked with identifying aspects that are **present in one document but absent in the other**.\\

\textbf{\#\#\# **Your Task:**}\\
1. Identify Unique Aspects:
 \begin{itemize} \renewcommand\labelitemi{-}
  \item Read both documents carefully.  
  \item Identify unique aspects or themes that are **discussed in only one document** but **not mentioned or addressed in the other document either implicitly or explicitly**.
 \end{itemize}     

2. Extract Representative Sentences:
 \begin{itemize} \renewcommand\labelitemi{-}
\item For each unique aspect, extract relevant sentences or paragraphs from the document where the aspect is **present**.
\item Do **not** extract anything from the document where the aspect is absent.
 \end{itemize}    
3. Provide Explanation:
\begin{itemize} \renewcommand\labelitemi{-}
\item Justify why the aspect is considered unique to the document.
\item Explain why it is **not found** in the other document.\\ 
 \end{itemize} 
 
\textbf{\#\#\# **Expected Output Format:**}
Provide at most 4 pairs, 2 for document 1 and 2 for document 2.
Your response must be a **single valid JSON object** structured as follows:

\begin{verbatim}
```json
[
  {
    "aspect": "Identified unique aspect or topic",
    "sentence1": if sentences found in document 1, else " "
    "sentence2": if sentences found in document 2, else " "
    "reason": "Brief explanation of why this aspect is unique to this document and absent in the other."
  },
  ......
]
```
\end{verbatim}

\begin{verbatim}
Input Documents:
    {Document 1}
    {Document 2}
\end{verbatim}

}
\end{tcolorbox}
\\

\begin{tcolorbox}[
    colback=gray!5,
    colframe=blue!60!black,
    title=\textbf{Prompt Template for GPT-4o Based Document Similarity Evaluation},
    fonttitle=\bfseries,
    sharp corners,
    boxrule=0.8pt,
    width=\textwidth
]

{\ttfamily
You are an intelligent system that evaluates the semantic similarity between two documents.\\

\textbf{\#\#\# **Your Task:**}
\begin{itemize} \renewcommand\labelitemi{-}
    \item Read both documents in their entirety.
    \item Compare the documents based on their overall informational content, including primary topics, central claims, and conclusions.
    \item Judge similarity based on substance, not surface wording or writing style.\\
\end{itemize}

\textbf{\#\#\# **Similarity Labels:**}

\begin{itemize} \renewcommand\labelitemi{-}
    \item \textbf{Highly Similar}: Both documents discuss the same primary topic and express the same central claims or conclusions. The overlap in informational content is strong, even if phrasing differs.
    \item \textbf{Somewhat Similar}: Both documents discuss the same primary topic, but overlap only partially in their core claims or supporting information. One document may include additional details, context, or emphasis not present in the other.
    \item \textbf{Marginally Similar}: The documents share only a broad theme or event, but differ in their main focus, claims, or conclusions. Any overlap is weak, indirect, or high-level.
    \item \textbf{Not Similar}: The documents do not share a primary topic, central claim, or conclusion, and there is no meaningful informational overlap.\\
\end{itemize}

\textbf{\#\#\# **Scoring:**}
\begin{itemize} \renewcommand\labelitemi{-}
    \item Assign a similarity label from the similarity label set.
    \item Assign a similarity score between \texttt{0.0} and \texttt{1.0} (inclusive).
    \item The score must be independent of the similarity label.
    \item Even when the same label is assigned, use different scores to reflect finer-grained differences.
\end{itemize}

\textbf{\#\#\# **Expected Output Format:**}
\begin{itemize} \renewcommand\labelitemi{-}
    \item Output a single valid JSON object.
    \item Do \textbf{not} include explanations, comments, or any text outside the JSON.
    \item The JSON object must contain exactly two keys: \texttt{"label"} and \texttt{"score"}.
\end{itemize}
\begin{verbatim}
```json
{
  "label": "Highly Similar | Somewhat Similar | Marginally Similar | Not Similar",
  "score": 0.0
}
```
\end{verbatim}
\begin{verbatim}
Input Documents:
Document 1:
{document1}

Document 2:
{document2}
\end{verbatim}

}
\end{tcolorbox}\\


\end{longtable}

\section{Prompts for \textsc{AspectSim} Variation and Baseline}
\label{appendix:design_prompts}

\scriptsize
\setlength{\tabcolsep}{0pt}
\renewcommand{\arraystretch}{1}

\begin{longtable}{p{\textwidth}}

\begin{tcolorbox}[
  breakable,
  enhanced,
  colback=gray!5,
  colframe=blue!60!black,
  title=\textbf{Aspect-Based Single-Sentence Extraction Prompt},
  fonttitle=\bfseries,
  sharp corners,
  boxrule=0.8pt
]
\ttfamily
You are a highly intelligent system designed to identify and extract a relevant sentence from a document based on a specific aspect.
You will first determine whether the document discusses the given aspect,
and then extract the \textbf{**most relevant sentence**} that directly discusses it.

\textbf{\#\#\# **To complete your task, follow these steps carefully:**}\\

\textbf{\#\#\# **Step 1: Aspect Presence Check**}
Does the document discuss the aspect \texttt{\{aspect\}}?
Answer \texttt{Yes} or \texttt{No}.
If \texttt{Yes}, proceed to Step~2.
If \texttt{No}, return an empty list for \texttt{narrative}.\\

\textbf{\#\#\# **Step 2: Sentence Extraction**}
If the answer in Step~1 is \texttt{Yes}, extract the
\textbf{single most relevant key sentence}
from the document that directly discusses the given aspect.\\

\textbf{\#\#\# **Output Constraints**}
Your output must follow these strict rules:
\begin{itemize}
  \item Output a single, valid JSON object.
  \item The JSON object must contain exactly two keys:
        \texttt{discusses\_aspect} and \texttt{narrative}.
  \item The value of \texttt{discusses\_aspect} must be either
        \texttt{Yes} or \texttt{No}.
  \item If \texttt{discusses\_aspect} is \texttt{Yes},
        \texttt{narrative} must contain exactly one sentence.
  \item If \texttt{discusses\_aspect} is \texttt{No},
        \texttt{narrative} must be an empty list.\\
\end{itemize}

\textbf{\#\#\# **Expected Output Format**}

\begin{verbatim}

{ "discusses_aspect": "Yes", "narrative": ["Matching sentence."] }
\end{verbatim}
\textbf{or}
\begin{verbatim}
{ "discusses_aspect": "No", "narrative": [] }

\end{verbatim}

\textbf{\#\#\# **Additional Instructions**}
\begin{itemize}
  \item Do not include explanations, comments, or any text outside the JSON object.
  \item Ensure that the extracted sentence is complete and well-formed.
  \item Do not provide partial or malformed JSON responses.
\end{itemize}

\textbf{\#\#\# **Final Instruction**}
Now analyze the following document: \texttt{\{document\}}, and determine whether it discusses the aspect: \texttt{\{aspect\}}.
If it does, extract the most relevant sentence that discusses the aspect.
\end{tcolorbox}

\\

\begin{tcolorbox}[
  breakable,
  enhanced,
  colback=gray!5,
  colframe=blue!60!black,
  title=\textbf{Aspect-Based Multi-Sentence Extraction Prompt},
  fonttitle=\bfseries,
  sharp corners,
  boxrule=0.8pt
]
\ttfamily
You are a highly intelligent system designed to identify and extract relevant sentences from a document based on a specific aspect.
You will first determine whether the document discusses the given aspect, and then extract all relevant sentences that discuss the aspect.

 \textbf{\#\#\# To complete your task, follow these steps carefully:}\\

 \textbf{\#\#\# **Step 1: Aspect Presence Check**} 
Does the document discuss the aspect \texttt{\{aspect\}}?
Answer \texttt{Yes} or \texttt{No}.
If \texttt{Yes}, proceed to Step~2.
If \texttt{No}, return an empty list for \texttt{narrative}.\\

 \textbf{\#\#\# **Step 2: Sentence Extraction**} 
If the answer in Step~1 is \texttt{Yes}, extract
\textbf{**all relevant sentences**} from the document that discusses the given aspect.\\

\textbf{\#\#\# **Output Constraints**}
Your output must follow these strict rules:
\begin{itemize}
  \item Output a single, valid JSON object.
  \item The JSON object must contain exactly two keys:
        \texttt{discusses\_aspect} and \texttt{narrative}.
  \item The value of \texttt{discusses\_aspect} must be either
        \texttt{Yes} or \texttt{No}.
  \item If \texttt{discusses\_aspect} is \texttt{Yes},
        \texttt{narrative} must be a list of sentences that
        discuss the aspect.
  \item If \texttt{discusses\_aspect} is \texttt{No},
        \texttt{narrative} must be an empty list.\\
\end{itemize}

 \textbf{\#\#\# **Expected Output Format**}

\begin{verbatim}

{ "discusses_aspect": "Yes", "narrative": ["Matching sentence 1.", "Matching sentence 2."] }
\end{verbatim}
\textbf{or}
\begin{verbatim}
{ "discusses_aspect": "No", "narrative": [] }

\end{verbatim}

\textbf{\#\#\# **Additional Instructions**}
\begin{itemize}
  \item Do not include any explanations, comments, or text outside the JSON object.
  \item Ensure that all extracted sentences are complete and well-formed.
  \item Do not provide partial or malformed JSON responses.
\end{itemize}

 \textbf{\#\#\# **Final Instruction**} \\
Now analyze the following document: \texttt{\{document\}}, and determine whether it discusses the aspect: \texttt{\{aspect\}}.
If it does, extract all relevant sentences that discuss the aspect.
\end{tcolorbox}

\\

\begin{tcolorbox}[
  breakable,
  enhanced,
  colback=gray!5,
  colframe=blue!60!black,
  title=\textbf{Aspect-Based Document Summarization Prompt},
  fonttitle=\bfseries,
  sharp corners,
  boxrule=0.8pt
]
\ttfamily
You are a highly intelligent system designed to identify and summarize information in a document that is relevant to a specific aspect. You will first determine whether the document discusses the given aspect, and then generate a concise summary that captures how the document addresses that aspect.

\#\#\# \textbf{To complete your task, follow these steps carefully:}\\

\textbf{\#\#\# **Step 1: Aspect Presence Check**} 
Does the document discuss the aspect \texttt{\{aspect\}}?
Answer \texttt{Yes} or \texttt{No}. \\
If \texttt{Yes}, proceed to Step~2. \\
If \texttt{No}, return an empty string for \texttt{summary}.\\

 \textbf{\#\#\# **Step 2: Aspect-Based Summarization**} 
If the answer in Step~1 is \texttt{Yes}, write a concise
\textbf{aspect-focused summary (2--4 sentences)}
describing how the document discusses the aspect.
The summary must focus only on information directly related to the aspect and ignore unrelated details.\\

\textbf{\#\#\# **Output Constraints**}
Your output must follow these strict rules:
\begin{itemize}
  \item Output a single, valid JSON object.
  \item The JSON object must contain exactly two keys:
        \texttt{discusses\_aspect} and \texttt{summary}.
  \item The value of \texttt{discusses\_aspect} must be either
        \texttt{Yes} or \texttt{No}.
  \item If \texttt{discusses\_aspect} is \texttt{Yes},
        \texttt{summary} must be the
        aspect-specific summary text.
  \item If \texttt{discusses\_aspect} is \texttt{No},
        \texttt{summary} must be an empty list.\\
\end{itemize}

\textbf{\#\#\#**Expected Output Format**}

\begin{verbatim}
{"discusses_aspect": "Yes", "summary": "Aspect-specific summary here."}
\end{verbatim}
\textbf{or}
\begin{verbatim}
{"discusses_aspect": "No", "summary": ""}
\end{verbatim}

\textbf{\#\#\# **Additional Instructions**}
\begin{itemize}
  \item Do not include any explanations, comments, or text outside the JSON object.
  \item Ensure that all summary sentences are complete, well-formed, and aspect-focused.
  \item Do not provide partial or malformed JSON responses.
  \item Keep the summary concise and factual (2--4 sentences maximum).
\end{itemize}

\textbf{\#\#\# **Final Instruction**} \\
Now analyze the following document: \texttt{\{document\}}, and determine whether
it discusses the aspect: \texttt{\{aspect\}}.
If it does, generate the aspect-based summary as described above.

\end{tcolorbox}\\

\begin{tcolorbox}[
  breakable,
  enhanced,
  colback=gray!5,
  colframe=blue!60!black,
  title=\textbf{Aspect-Based Document Similarity Scoring Prompt (LBS Baseline)},
  fonttitle=\bfseries,
  sharp corners,
  boxrule=0.8pt
]
\ttfamily
You are a highly intelligent system designed to measure the similarity between two documents on a scale of 0 to 1 with respect to a specific aspect. You will first determine whether the documents discuss the given aspect explicitly or implicitly, and then provide a similarity score corresponding to that aspect.

\textbf{\#\#\# **To complete your task, follow these steps carefully:**}\\

\textbf{\#\#\# **Step 1: Read the Documents**}
Read both documents thoroughly. \\

\textbf{\#\#\# **Step 2: Aspect Presence Check**} 
Determine whether \textbf{both documents} discuss the aspect
\texttt{\{aspect\}}. Answer \texttt{Yes} or \texttt{No}.
\begin{itemize}
  \item If \texttt{No}, return a similarity score of \texttt{0}.
  \item If \texttt{Yes}, proceed to Step~3.\\
\end{itemize}

\textbf{\#\#\# **Step 3: Aspect-Based Similarity Scoring**}
If the answer in Step~2 is \texttt{Yes}, determine the similarity of the sentences or content in the two documents that discuss the given aspect. Assign a score and label according to the following guidelines:
\begin{itemize}
  \item \textbf{Marginally Similar}: assign a score between \texttt{0.1} and \texttt{0.4}.
  \item \textbf{Somewhat Similar}: assign a score between \texttt{0.5} and \texttt{0.7}.
  \item \textbf{Highly Similar}: assign a score between \texttt{0.7} and \texttt{1.0}. \\
\end{itemize}

\textbf{\#\#\# **Output Constraints**}
Your output must follow these strict rules:
\begin{itemize}
  \item Output a single, valid JSON object.
  \item The JSON object must contain exactly three keys:
        \texttt{discusses\_aspect}, \texttt{label} and \texttt{score}.
  \item The value of \texttt{discusses\_aspect} must be either
        \texttt{Yes} or \texttt{No}.
  \item The value of \texttt{score} must always be a real number
        (e.g., \texttt{0.2}) representing the similarity score.
\end{itemize}

\#\#\# \textbf{Expected Output Format}

\begin{verbatim}
{ "discusses_aspect": "Yes", "label": <similarity label>,  "score": 0.7 }
\end{verbatim}
\textbf{or}
\begin{verbatim}
{ "discusses_aspect": "No", "label": Not-found, "score": 0 }
\end{verbatim}

\textbf{\#\#\# **Additional Instructions**}
\begin{itemize}
  \item Do not include any explanations, comments, or text outside the JSON object.
  \item Think carefully before assigning a similarity label and score, and ensure it reflects the true similarity with respect to the aspect.
  \item Do not provide partial or malformed JSON responses.
\end{itemize}

 \textbf{\#\#\# **Final Instruction**} \\
Now analyze the following documents:
\texttt{\{document1\}} and \texttt{\{document2\}}, and determine their similarity with respect to the aspect: \texttt{\{aspect\}}.
\end{tcolorbox}

\end{longtable}

\section{Data Samples}
\label{appendix:data-sample}

\tiny
\setlength{\tabcolsep}{6pt}
\renewcommand{\arraystretch}{1.1}

\begin{longtable}{p{0.12\textwidth} p{0.84\textwidth}}

\toprule
\textbf{Field} & \textbf{Content} \\
\midrule
\endfirsthead

\toprule
\textbf{Field} & \textbf{Content (continued)} \\
\midrule
\endhead

\midrule
\multicolumn{2}{r}{\textit{Continued on next page}} \\
\endfoot

\endlastfoot

\multicolumn{2}{c}{\textbf{Domain: Peer Review}} \\
\midrule

Document 1 &
\textit{Summary Of The Paper.}
\HighSimText{This paper discusses solving robust subspace recovery by solving parallel versions of st , for random initializations of , using projected subgradient method.} \SomeSimText{The argument is that using this method, in the limit where the number of inliers greatly outnumber the outliers, that if the inlier dimension is c, then c random initializations will converge on identifying this dimension with probability 1.} Experimental results and proofs are given to prove this idea.

\textit{Main Review.}
The idea is interesting and seems to be more of a statistical result than an optimization. The numerical results are encouraging, and I cant directly find any issues with the proofs though for time reasons, I did not look too closely. \MargSimText{There are two assumptions I think should be elaborated, at least qualitatively first, the condition 9 on the initialization of bs, and second, the overestimate of c over c.} These relationships would really give better intuition as to how powerful the random initialization is. Additionally, I am curious as to how much it matters that the data itself has isometry principles. I would suspect if the inlier dimensions have a very uneven contribution, this method should not work that well how is this captured

\textit{Summary Of The Review.}
Overall, I think the paper's idea is nice and looks sound; however, I am not an expert in this area.
\\
\addlinespace

Document 2 &
\HighSimText{The work proposed a simple framework that allows performing robust subspace recovery without requiring a priori knowledge of the subspace codimension.} \SomeSimText{The proposed approach is based on Dual Principal Component Pursuit DPCP, which is amenable to handling subspaces of high relative dimensions.} Empirical results corroborate the developed theory and showcase the merits of the proposed optimization methods.

\textit{Main Review.}
The work proposed a new analysis framework for justifying the robust subspace recovery under the DPCP formulation without requiring a priori knowledge of the subspace codimension. The work analyzed both the gradient flow and its projected subgradient counterpart, showing convergence to the target solutions. The paper is well-organized and well-presented overall. The work tackles an important challenge to perform robust subspace recovery in a high relative subspace dimension regime without requiring a priori knowledge of the true subspace dimension. \MargSimText{Nonetheless, the reviewer is a little concerned over the novelty of the work. It seems to be borrowing and combining ideas from previous work on DPCP and implicit bias for low-rank matrix factorization.}

\textit{Summary Of The Review.}
The work is well-presented and tackles the challenge of unknown true subspace dimension via overparametrization and implicit bias of the optimization method. The work combines ideas from previous work on DPCP and implicit bias for low-rank matrix recovery. The overall result is novel, while the approaches seem to be a combination of previous methods.

\\
\addlinespace

Aspect \& Similarity &
\textbf{Robust Subspace Recovery}: \HighSim{ \textit{Highly Similar}} (Both sentences discuss the topic of robust subspace recovery, focusing on methods that do not require prior knowledge of certain parameters. They both highlight the approach used to achieve this recovery.
) \newline
\textbf{Methodology and Approach}: \SomeSim{\textit{Somewhat Similar}} (Both sentences discuss the methodology for subspace recovery, but they focus on different aspects: one on the convergence of random initializations and the other on the DPCP approach. They share a common theme of handling high-dimensional subspaces.
) \newline
\textbf{Concerns and Assumptions}: \MargSim{\textit{Marginally Similar}} (The first sentence discusses specific assumptions that need elaboration, while the second sentence expresses concerns about the novelty of the work. They both express concerns but focus on different issues.
)
\\

\midrule
\multicolumn{2}{c}{\textbf{Domain: User Opinion}} \\
\midrule

Document 1 &
This is a decent place to stay and will leave you feeling happy overall. Initially, there was a bit of a rough start here at the hotel, but it's still a nice hotel. \MargSimText{There is some peace and quiet here at this hotel, but it is still quite close to the action with restaurants and shopping nearby.} \SomeSimText{The room was clean with adequate disabled access , but it may take a while to get access to the room.} \HighSimText{The staff here is okay and mostly very helpful.} They do enforce valet parking, however, which is a little off-putting. Dinner at this hotel was really good and pleasing. There are also other food options nearby, such as a restaurant that serves Italian food. There are valet and parking services available, too.
\\
\addlinespace

Document 2 &
The hotel has a deal that you get the fourth night free if you stay three nights. The hotel is in close proximity to a zoo and aquarium, amongst other tourist sights. The hotel room overlooked the city, Diamond Head, and the zoo. \SomeSimText{The room and bathroom were an adequate size, but could generally do with a deep clean.} \HighSimText{The staff at the hotel were friendly and knowledgeable for the most part, although there were some who were not as helpful.} There were free buffets every single day at this hotel, which has to be taken advantage of. \MargSimText{There was a small outdoor patio, which unfortunately overlooked the pool and other patios, and that made it quite noisy.} The tiles around the hotel's swimming pool are very slippery, so caution is advised if using the facility.
\\
\addlinespace

Aspect \& Similarity &
\textbf{Staff and Service
}: \HighSim{ \textit{Highly Similar}} (Both sentences describe the staff as generally helpful and friendly, with some exceptions. The core idea about the staff's helpfulness is consistent.
) \newline
\textbf{Room and Cleanliness
}: \SomeSim{\textit{Somewhat Similar}} (Both sentences discuss the room's condition, but while one mentions cleanliness and access, the other suggests a need for a deep clean. They overlap on the topic of room condition but differ in details.
) \newline
\textbf{Noise Levels
}: \MargSim{\textit{Marginally Similar}} (Both sentences discuss noise conditions, but while one suggests general peace with nearby activity, the other describes a specific area that was explicitly noisy; they overlap on noise level, but they differ both in intensity and context.) \newline
\\

\midrule
\multicolumn{2}{c}{\textbf{Domain: Medical Literature}} \\
\midrule

Document 1 &
\textit{Background}---Hypertension is a leading mortality risk factor yet inadequately controlled in most affected subjects. Effective programs to address this problem are lacking. We hypothesized that an information technology-supported management program could help improve blood pressure (BP) control.
\textit{Methods and Results}---This randomized controlled trial included 223 primary care hypertensive subjects with mean 24-hour BP $>$130/80 and daytime BP $>$135/85 mm Hg measured with ambulatory monitoring (ABPM). \SomeSimText{Intervention subjects received a BP monitor and access to an information technology-supported adherence and BP monitoring system providing nurses, pharmacists, and physicians with monthly reports.} Control subjects received usual care. The mean ($\pm$ SD) follow-up was 348 ($\pm$78) and 349 ($\pm$84) days in the intervention and control group, respectively. The primary end point of the change in the mean 24-hour ambulatory BP was consistently greater in intervention subjects for both systolic (-11.9 versus -7.1 mm Hg; $P<0.001$) and diastolic BP (-6.6 versus -4.5 mm Hg; $P=0.007$). The proportion of subjects that achieved Canadian Guideline target BP (46.0\% versus 28.6\%) was also greater in the intervention group ($P=0.006$). We observed similar BP declines for ABPM and self-recorded home BP, suggesting the latter could be an alternative for confirming BP control. \MargSimText{The intervention was associated with more physician-driven antihypertensive dose adjustments or changes in agents ($P=0.03$), more antihypertensive classes at study end ($P=0.007$), and a trend toward improved adherence measured by prescription refills ($P=0.07$).}
\textit{Conclusions}---\HighSimText{This multidisciplinary information technology--supported program that provided feedback to patients and healthcare providers significantly improved blood pressure levels in a primary care setting.}

\\
\addlinespace

Document 2 &
\textit{Background}---Lifestyle modification is recommended for management of prehypertension, yet finding effective interventions to reach rural women is a public health challenge. \SomeSimText{This community-based clinical trial compared the effectiveness of standard advice to two multi-component theory-based tailored interventions, using web-based or print-mailed delivery, in reducing blood pressure among rural women, ages 40--69, with prehypertension.}
\textit{Methods}---A total of 289 women with prehypertension enrolled in the Wellness for Women: DASHing Towards Health trial, a 12-month intervention with 12-month follow-up. Women were randomly assigned to groups using a 1:2:2 ratio, comparing standard advice (30-minute counseling session) to two interventions (two 2-hour counseling sessions, five phone goal-setting sessions, a strength-training video, and 16 tailored newsletters delivered via web-based or print-mailed formats). Linear mixed model methods were used to test planned pairwise comparisons of marginal mean change in blood pressure, healthy eating, and activity, adjusted for age and baseline level. Generalized estimating equations were used to examine the proportion of women achieving normotensive status and meeting health outcome criteria for eating and activity.
\textit{Results}---Mean blood pressure reduction ranged from 3.8 (SD = 9.8) mm Hg to 8.1 (SD = 10.4) mm Hg. The 24-month estimated marginal proportions of women achieving normotensive status were 47\% for the web-based group and 39\% for both the print-mailed and standard advice groups, with no significant group differences ($p = 0.11$ and $p = 0.09$, respectively). \MargSimText{Web-based and print-mailed groups improved more than the standard advice group for waist circumference ($p = 0.017$ and $p = 0.016$, respectively), percentage of daily calories from fat ($p = 0.018$ and $p = 0.030$) and saturated fat ($p = 0.049$ and $p = 0.013$), daily servings of fruits and vegetables ($p = 0.008$ and $p < 0.005$), and low-fat dairy intake ($p < 0.001$ and $p = 0.002$).} Greater improvements were observed in web-based versus standard advice groups in systolic blood pressure ($p = 0.048$) and estimated VO$_2$max ($p = 0.037$). Dropout rates were 6\% by 6 months and 11.4\% by 24 months, with no differences across groups.
\textit{Conclusions}---\HighSimText{Rural women with prehypertension receiving distance-delivered, theory-based lifestyle interventions can achieve reductions in blood pressure and attain normotensive status.}
\\
\addlinespace

Aspect \& Similarity &
\textbf{Conclusions on Blood Pressure Management}: \HighSim{ \textit{Highly Similar}} (Both sentences highlight the effectiveness of interventions in managing blood pressure.
) \newline
\textbf{Intervention Methods}: \SomeSim{\textit{Somewhat Similar}} (In both sentences, the idea of intervention of blood pressure management is similar, but the methods differ.) \newline
\textbf{Adherence and Lifestyle Changes}: \MargSim{\textit{Marginally Similar}} (Although both relate to adherence and lifestyle, the specific focus and outcomes are different.
) \newline
\\

\midrule
\multicolumn{2}{c}{\textbf{Domain: Factual Event News}} \\
\midrule

Document 1 &
ALGIERS Reuters  President Abdelaziz Bouteflika landed in Algeria on Sunday Ennahar TV said, after spending two weeks in a Swiss hospital while protesters have massed at home in the biggest threat to his 20year rule. An Algerian government plane transporting the president arrived at Boufarik military airport just southwest of the capital. Top Algerian officials often use that airport. Tens of thousands of Algerians of all social classes have protested over the past three weeks against Bouteflikas decision to stand in Aprils election. In the clearest indication yet that the generals sympathize with them, the chief of staff said the military and the people had a united vision of the future, state TV reported. Lieutenant General Gaed Salah did not mention the protests. \HighSimText{The ruling FLN party urged all sides to work together to end the political crisis, Ennahar TV said. It wants national reconciliation and to preserve security and stability, the station said.} But there are no signs Algerians are prepared to heed that call after rejecting Bouteflikas offer to limit his term after the election. Bouteflikas system is over, said a commentator on Ennahar TV, which is close to the presidents inner circle, noting that some of the Bouteflikas allies have joined the protests. Ennahar TV also said Bouteflika underwent routine medical checks in Geneva. On Sunday, thousands took to the streets of the capital carrying the Algerian flag and chanting Bouteflika, there will be no fifth term. Many shops in Algiers were closed and residents said train services had been suspended. We have taken to the streets today to protest a fifth presidential term. We are against a fifth term. This is enough, protester Zakaria said in front of the Central Postal Office. \SomeSimText{Young Algerians are desperate for jobs and angry about unemployment, corruption and an elderly elite in the North African country, a major oil and gas producer.} The current system is unable to provide jobs, said Farid Kahil, 27, who is unemployed. A general strike has hit the countrys Mediterranean oil ports of Skikda and Bejaia but exports were not affected, according to port staff. Bouteflika managed to remain in power as the Arab Spring uprising toppled autocrats in neighboring countries in 2011 because Algeria had enough foreign reserves to boost state spending. Older Algerians haunted by the civil war in the 1990s tolerated crackdowns on dissent in exchange for stability, giving the government some breathing space. Now some have appeared at demonstrations to demand reforms. We need a new generation to govern us and secure a better future for our children, said pensioner Ahmed, 63. Even if Bouteflikas position becomes untenable, it is not clear who could replace him. Algeria has stagnated for decades under veterans of the independence war who dominate the country. For years, rumors have swirled about potential successors, but no one credible has emerged who has the backing of the army and elite and is not in their 70s or 80s. Algerias chief of staff has warned that chaos would not be tolerated. The military has stayed in barracks. Several public figures, including members of Bouteflikas FLN party and lawmakers have resigned to join the rallies against a political system dominated by war veterans since independence from France in 1962. Two branches of the powerful labor union UGTA, representing tens of thousands of workers, have opposed the reelection plan and lawyers have also joined rallies. Algerians packed central Algiers to capacity on Friday in the biggest protests in the capital in 28 years. Security forces detained 195 people, state television said, citing offences such as looting. 
\\
\addlinespace

Document 2 &
ALGIERS Reuters  More than 1,000 judges said they would refuse to oversee Algerias election if President Abdelaziz Bouteflika contests it, and clerics defied any state role in their work, in a double rebuff to an ailing leader fighting for his political survival. Bouteflika, who returned to Algeria on Sunday after medical treatment in Switzerland, has watched one longtime ally after another join mass demonstrations calling on him to step down. \SomeSimText{Now in their third week, the protests have seen Algerians desperate for jobs and angry about unemployment and corruption demonstrate in towns across around the vast North African country against Bouteflikas bid for a fifth term in office.} The marches have shattered years of political inertia and unsettled Algerias opaque but powerful security establishment. In a statement, the judges added their voice to the protests by announcing the formation of a new association to restore the gift of justice. We announce our intention to abstain from ... supervising the election process against the will of the people, which is the only source of power, the judges said in a statement. The statement drew a sharp retort from Algerian Justice Minister Tayeb Louh, a member of Bouteflikas inner circle, who said judges should remain neutral. The independence and integrity of the judge must be consistent whatever the reasons, Ennahar TV quoted him as saying. In another setback for the president, who plans to stand in elections in April, clerics told the minister of religious affairs to stop pressuring them to issue progovernment sermons. Leave us to do our job, do not interfere, cleric Imam Djamel Ghoul, leader of an independent group of clerics, said in remarks to reporters. The 82yearold Bouteflika faces the toughest fight of his 20yearold rule, following a tenure in which he became the north African countrys most powerful president in 30 years. The secretive militarybased establishment known to Algerians as le pouvoir the powersthatbe appears to have stood aside while the demonstrations have taken place. Bouteflika is back, we delivered a message, we need a response, and we need a response now, pharmacist Mouloud Mohamed, 29, told Reuters. In Algiers, tens of unionists staged a protest rally outside the headquarters of the main union, UGTA, calling on its leader Abdelmadjid Sidi Said, a Bouteflika ally, to resign. The veteran head of state has rarely been seen in public since a stroke in 2013. Last April, he appeared in Algiers in a wheelchair. In the clearest indication yet that the army is seeking to put some distance between itself and Bouteflika, the chief of staff said the military and the people had a united vision of the future, state TV reported. Lieutenant General Gaed Salah did not mention the unrest. \HighSimText{His ruling FLN party urged all sides to work together to end the crisis and promote national reconciliation, Ennahar TV said. But some of its members have quit and joined ranks with demonstrators.} Even if Bouteflika is forced from office, there is no clear replacement, raising the strong possibility that the ruling elite will maintain their grip. Political sources say that if he caves in to the demonstrators demands, several parties including the military, war veterans and members of the opposition would need to build consensus on a way forward to secure an orderly transition. The process could also include prominent protesters. For years, rumors have swirled about potential successors, but no one credible has emerged who has the backing of the army and the elite and is not at least 70. The end of the reign of the outgoing president is over. It is just a matter of time, said Louisa Ait Idriss, a university professor and prominent political analyst. The challenge is to end Bouteflika without having another Bouteflika. The protests come four years after Bouteflika consolidated his position by dismissing military intelligence chief Mohamed Mediene, a rival once seen as Algerias Kingmaker, a move many expected would allow the president to step aside for an ally. Bouteflikas dismissal of the general was the culmination of a struggle to impose his authority on military intelligence, a leading player in the civil war of the 1990s, and make the presidency the true center of national power.
\\
\addlinespace

Aspect \& Similarity &
\textbf{FLN Party's Call for Reconciliation}: \HighSim{ \textit{Highly Similar}} (Both sentences convey the FLN party's call for national reconciliation and cooperation to resolve the political crisis.) \newline
\textbf{Economic Concerns and Unemployment}: \SomeSim{\textit{Somewhat Similar}} (Both sentences highlight economic concerns, such as unemployment and corruption, as reasons for the protests, but the second sentence ties these issues directly to the protests against Bouteflika's fifth term.) \newline
\\
\midrule
\multicolumn{2}{c}{\textbf{Domain: Political News}} \\
\midrule

Document 1 &
Capitol police officers stand guard as protesters gather on the U.S. Capitol Building in Washington on Jan. 6, 2021. \HighSimText{Capitol Police Rejected Multiple Offers of Federal Help. U.S. Capitol Police  USCP  rejected offers of federal manpower days before Jan. 6 and they again turned down an offer of federal reinforcements as protestors breached the Capitol building during the formal counting of electoral votes.} \SomeSimText{Three days before Jan. 6, the Pentagon asked Capitol Police if they needed National Guard manpower, but their offer was turned down by Capitol Police, unnamed senior defense officials told the Associated Press. And as the building was being stormed, Justice Department leaders reached out to Capitol police and offered to provide them with FBI agents. But Capitol Police again turned them down, according to the sources.} The Epoch Times reached out to USCP, the Department of Defense, and the Justice Department for comment, none immediately responded. The rejections for additional federal support came despite plenty of warnings, ample resources, and time to prepare. USCP had only planned for a free speech demonstration. Four protesters died, including one shot inside the building. Ashli Babbitt was an Air Force veteran who was shot and killed by Capitol Police as protestors swarmed into the building. \HighSimText{Army Secretary Ryan McCarthy said it became clear that the Capitol Police were overrun.} He said there was no contingency planning done in advance for what forces should do in case of a problem at the Capitol because Defense Department help was turned down.  They ve got to ask us, the request has to come to us,  said McCarthy. U.S. Capitol Police Chief Steven Sund, under pressure from Schumer, Pelosi, and other congressional leaders, was forced to resign. Senate Majority Leader Mitch McConnell asked for and received the resignation of the Sergeant at Arms of the Senate, Michael Stenger, effective immediately. Paul Irving, the longtime Sergeant at Arms of the House, also resigned. It is not clear how many officers were on duty Wednesday, but the complex is policed by a total of 2,300 officers over 16 acres of ground, who protect the 435 House representatives, 100 U.S. senators, and their staff. The breach of the capitol halted a joint session, interrupting concurrent debates in the House and Senate over an objection to the counting of a slate of presidential electors from Arizona for President elect Joe Biden.
Washington Mayor Muriel Bowser declared a citywide curfew from 6 p.m. to 6 a.m after the reports of the breach. \MargSimText{Trump issued several calls for peace as the events unfolded. Please support our Capitol Police and Law Enforcement. They are truly on the side of our Country. Stay peaceful, the president wrote on Twitter}. Earlier in the day, tens of thousands of people gathered at a rally near the White House on Jan. 6 to hear from the president at a pivotal moment and to voice their concerns about what they see as a fraudulent election. Supporters of President Donald Trump lined the entire area surrounding The Ellipse, a 52-acre park located just south of the White House fence, with many traveling by car or plane from around the country. Trump has not conceded the race but has said there will be a peaceful transition of power. Congress certified Biden as the winner early on Thursday morning. Biden said in a Jan. 7 statement that the breach was not dissent, it was disorder. 
They weren't protestors, they were rioters, insurrectionists, and domestic terrorists,  Biden said. Trump, in a Twitter post on Friday, said he would not be attending the inauguration later this month. To all of those who have asked, I will not be going to the Inauguration on January 20th,  he said.

\\
\addlinespace

Document 2 &
Just hours after being the site of a historic siege, the scene outside of the U.S. Capitol was eerie but quiet. The crowds were gone, and the tear gas had lifted. The only sound on the streets was the occasional siren in the distance. From police cruisers to blacked out SUVs, nearly every vehicle on the road belonged to one law enforcement agency or another. A few dozen TV anchors filed reports from the Capitol's southwest side. Over their shoulders, camouflaged government employees of unclear affiliation ambled about, relocating their shields and other tactical gear. Closer to the building, a column of black clad riot cops shuffled through shadows, visible only by the glint of their helmets. The law enforcement presence Wednesday night, like the one used on protesters at Donald Trump s inauguration in 2017 and Black Lives Matter protesters in 2020, was a reminder that the capital of the world s most powerful country does not lack for security forces. As a collective body, the agencies are abundant and well armed, and they are tapped into a billion-dollar intelligence-sharing apparatus. Not that Wednesday s rally was some secret, nor was its conclusion unforeseeable. The rally that preceded the storming of the Capitol had been hyped for weeks, including by the president himself, who for the past four years has had a direct relationship to many acts of staggering right wing violence. \HighSimText{It definitely shows the ineffectiveness of the intelligence network that we've built since 9 11, that the Capitol Police would not have been prepared for an assault on the Capitol that was planned in public,  Mike German, a fellow at the Brennan Center for Justice and former FBI agent specializing in counterterrorism, told The Intercept.}  It wasn t as if this was a spontaneous gathering. This was an event that had been planned for weeks, and it was very clear in the social media activity and public statements of these militants what they intended to do.  The onslaught came as Congress met to certify Electoral College votes for President elect Joe Biden. In response, the president's supporters organized a  Stop the Steal rally that drew thousands of demonstrators into the heart of Washington, D.C. It was just before 1 p.m. when the crowd, with the president s encouragement, began toppling barricades and pushing their way past Capitol Police. Rep. Chris Pappas, a New Hampshire Democrat, who was among the lawmakers rushed from the building, described the speed with which the crowd overwhelmed the Capitol Police as  breathtaking.  \SomeSimText{In a statement released Thursday, Capitol Police Chief Steven A. Sund, said his officers  responded valiantly when faced with thousands of individuals involved in violent riotous actions,  and noted that nearly 20 state, local, and federal agencies, including the National Guard, responded to the days events}. More than 50 officers were injured, he said, adding,  The violent attack on the U.S. Capitol was unlike any I have experienced in my 30 years in law enforcement here in Washington D.C.  According to reports Thursday, the Metropolitan Police Department had  no intelligence  suggesting  there would be a breach of the US Capitol.  German, the former FBI agent, said,  There needs to be a serious inquiry into the intelligence breakdowns and the tactical breakdowns that allowed the capital itself to be breached, particularly by people with weapons.  Adam Isacson, the D.C. based director of the defense oversight program at the Washington Office on Latin America, linked the events to a broader politicization of law enforcement under Trump, reminiscent of the anti democratic movements the U.S. has historically sponsored in countries around the world.  You don't get to ransack the Capitol for hours, then calmly walk away, unless law enforcement and its command share your views,  he wrote.  What we saw yesterday was tacit approval of the rioters. Full stop.  For those who have followed far-right violence during the Trump presidency, the most shocking thing about the events of January 6, 2021, may be just how predictable it all was. \MargSimText{In an address before the violence began, the president, who had long made clear that he had no interest in conceding the election, reminded his supporters that they had been wronged by evil actors and that it was up to them to fight back for the fate of the country. He anointed them the stars of their own action movie, with the final climactic scene now at hand.  If you don t fight like hell,  Trump warned,  you re not going to have a country anymore.}  He told his followers that together they would march on the Capitol and make their voices heard. Of course, he personally did not do this. Trump climbed into an armored vehicle and left the scene, but his supporters did precisely what they were told to do. \HighSimText{The insurrectionary mob was met with some resistance as they descended on the Capitol some, but not much}. Police did use chemical agents against the crowd, but by and large the response bore little resemblance to the iron fisted crackdown on racial justice protesters witnessed in Washington, D.C., and cities across the country just months before. The rampagers shouted at the police, calling them traitors, and threw punches at officers without consequence. ...................<continued>
\\
\addlinespace

Aspect \& Similarity &
\textbf{Capitol Police's preparedness and response}: \HighSim{ \textit{Highly Similar}} (Both sentences discuss the lack of preparedness and inadequate response of the Capitol Police during the breach.) \newline
\textbf{Federal offers of assistance}: \SomeSim{\textit{Somewhat Similar}} (Both sentences context is federal assistance, but the details differ.) \newline
\textbf{Trump's role and statements}: \MargSim{\textit{Marginally Similar}} (Document 1 presents Trump's calls for peace, while Document 2 highlights his incitement and encouragement of the crowd. The perspectives on Trump's role are conflicting.) \newline
\\
\bottomrule
\caption{Example instances from all five domains in the dataset. Each domain includes a document pair, a few aspects, and their corresponding similarity labels.}
\label{table:multi-domain-examples} 

\end{longtable}

\label{sec:appendix}

\end{document}